\batchmode
\makeatletter
\def\input@path{{\string"F:/Trabajo laptop/Mis articulos/Finished/Set of trajectories framework/Accepted/\string"}}
\makeatother
\documentclass[english]{IEEEtran}
\usepackage[latin9]{inputenc}
\usepackage{float}
\usepackage{textcomp}
\usepackage{amsmath}
\usepackage{amsthm}
\usepackage{amssymb}
\usepackage{stmaryrd}
\usepackage{graphicx}
\usepackage{nomencl}
\providecommand{\printnomenclature}{\printglossary}
\providecommand{\makenomenclature}{\makeglossary}
\makenomenclature

\makeatletter

\newcommand{\lyxmathsym}[1]{\ifmmode\begingroup\def\b@ld{bold}
  \text{\ifx\math@version\b@ld\bfseries\fi#1}\endgroup\else#1\fi}

\providecommand{\tabularnewline}{\\}
\floatstyle{ruled}
\newfloat{algorithm}{tbp}{loa}
\providecommand{\algorithmname}{Algorithm}
\floatname{algorithm}{\protect\algorithmname}

\theoremstyle{plain}
\newtheorem{thm}{\protect\theoremname}
\theoremstyle{definition}
\newtheorem{example}[thm]{\protect\examplename}
\theoremstyle{plain}
\newtheorem{lem}[thm]{\protect\lemmaname}

\usepackage{cite} 
\usepackage[margin=8pt,font=footnotesize]{caption}
\allowdisplaybreaks 
\usepackage{algorithm}
\usepackage{algpseudocode}

\floatname{algorithm}{Procedure} 

\@ifundefined{showcaptionsetup}{}{%
 \PassOptionsToPackage{caption=false}{subfig}}
\usepackage{subfig}
\makeatother

\usepackage{babel}
\providecommand{\examplename}{Example}
\providecommand{\lemmaname}{Lemma}
\providecommand{\theoremname}{Theorem}

\begin{document}

\title{Multiple target tracking based on sets of trajectories}

\author{Ángel F. García-Fernández, Lennart Svensson, Mark R. Morelande\thanks{A. F. García-Fernández is with the Department of Electrical Engineering and Electronics, University of Liverpool, Liverpool L69 3GJ, United Kingdom, and also with the ARIES research centre, Universidad Antonio de Nebrija, Madrid, Spain, (email: angel.garcia-fernandez@liverpool.ac.uk). L. Svensson is with the Department of Electrical Engineering, Chalmers University of Technology, SE-412 96 Gothenburg, Sweden (email: lennart.svensson@chalmers.se). M. R. Morelande is with the National Australia Bank, 800 Bourke St., Melbourne 3000, Victoria, Australia (email: m.morelande@gmail.com).
The authors would like to thank Abu S. Rahmathullah, Jason L. Williams, Karl Granstr{ö}m, Raghavendra Selvan, Roy Streit and Yuxuan Xia for helpful comments.}}
\maketitle
\begin{abstract}
We propose a solution of the multiple target tracking (MTT) problem
based on sets of trajectories and the random finite set framework.
A full Bayesian approach to MTT should characterise the distribution
of the trajectories given the measurements, as it contains all information
about the trajectories. We attain this by considering multi-object
density functions in which objects are trajectories. For the standard
tracking models, we also describe a conjugate family of multitrajectory
density functions. 
\end{abstract}


\section{Introduction}

\begin{figure}
\begin{centering}
\includegraphics{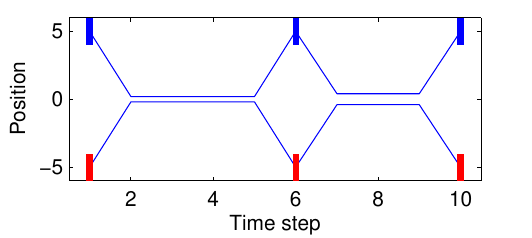}
\par\end{centering}
\caption{\label{fig:llustration-trajectory-questions}Illustration of a trajectory-related
question in one-dimensional case. Two targets are initially separated,
get in close proximity and separate twice, as represented by thin
blue lines. A possible trajectory-related question is, what is the
probability that the same target (plane) was in interval (city) $A=\left[4,6\right]$,
shown as a thick blue line, at time 6 and in interval $B=\left[-4,-6\right]$,
shown as a thick red line, at time 10? }

\end{figure}

Multiple target tracking (MTT) has an extensive range of applications,
for example, in surveillance \cite{Blackman04}, robotics \cite{Shulz01}
or computer vision \cite{Maggio08}. In MTT, sensors obtain noisy
measurements from targets that appear, move and disappear from a scene
of interest, forming trajectories or tracks. In MTT, we are interested
in answering target and trajectory-related questions that may arise
in the application under consideration. For example, what are the
best target estimates at the current time (according to a certain
criterion)? What are the best trajectory estimates? As illustrated
in Figure \ref{fig:llustration-trajectory-questions}, after two planes
fly around for some time, what is the probability that the same plane
was in city A at a certain time and is in city B now?

From a Bayesian perspective, after observing noisy measurements from
a random variable, all available knowledge about this random variable
is included in its conditional distribution given the measurements
\cite{Robert_book07}. Therefore, this distribution enables us to
answer all possible questions about the considered random variable.
In this paper, we are interested in how to represent this variable/state
in MTT and how to characterise its distribution so that we have a
full Bayesian solution of the MTT problem and can answer all types
of target and trajectory related questions. We focus on the MTT problem
with targets without a unique identification. That is, there is not
a unique way in which a particular target moves or affects measurements,
which is the common case in radar applications \cite{Reid79,Mori86,Blackman_book99,Mahler_book07}.
We proceed to review state representations used in the literature,
along with their pros and cons, before stating our contributions and
their practical implications. 

Original derivations of classic MTT algorithms, such as multiple hypothesis
tracking (MHT) \cite{Reid79} and joint probabilistic data association
(JPDA) \cite{Fortmann83} do not explicitly use a representation of
the multitarget state. Nevertheless, later papers on MHT and JPDA
represent the multitarget state at a certain time step as a vector
or a sequence \cite{Kurien_inbook90,Vermaak05,Mori86}. A set representation
\cite{Mahler_book07,Mahler_book14} in multi-object systems has some
advantages over vector/sequence representation: we avoid the arbitrary
ordering of the objects inherent in the multi-object state vector/sequence
and we can define mathematical metrics for algorithm evaluation and
estimator design \cite{Schuhmacher08,Guerriero10}. 

An appealing and rigorous way of dealing with such multiple object
systems from a Bayesian point of view is to use the random finite
set (RFS) framework and finite-set statistics (FISST) developed by
Mahler \cite{Mahler_book07,Mahler_book14}. In the proposed RFS algorithms
for MTT, the state at a certain time step is the (random) set of targets
at this time step and, as in classic approaches to MTT, the main focus
has been on the filtering problem. That is, we recursively calculate/approximate
the multitarget density of the current set of targets given current
and past measurements \cite{Mahler_book14}. This density is referred
to as the multitarget filtering density and contains all information
of interest about the targets at the current time. Based on the multitarget
filtering densities, we can answer target-related questions at the
current time step, such as target state estimation. However, we cannot
answer trajectory-related questions, such as the one illustrated in
Figure \ref{fig:llustration-trajectory-questions}, and it is not
obvious how to build trajectories in a sound manner.

The most popular approach to try to build trajectories from first
principles consists of adding a unique label to each single target
state so that each target is identified over its life time \cite{Angel09,Angel13,Vo13,Vo14,Ma06,Aoki12b,Aoki16,Craciun15}\cite[Sec. 14.5.6]{Mahler_book07},
though other approaches exist \cite{Houssineau18}. Labels can appear
in two forms. If a label is explicit, such as the registration number
of an aircraft or the name of a person, labels may have a physical
meaning and are referred to as target IDs. However, in many cases,
these target IDs are not observable \cite{Reid79,Mori86,Blackman_book99,Mahler_book07}
and there is total uncertainty about them. For example, it is not
possible to infer the serial number of a missile using radar measurements
and it is not usually of interest. 

When IDs cannot be inferred and therefore do not form part of the
model, as in this paper, one approach to build trajectories is to
add implicit labels, which we simply refer to as labels, rather than
the explicit labels. In this case, labels are unobservable, static
and uniquely assigned to targets when they are born following a certain
convention. In a deterministic setting, labels allow us to identify
trajectories from a sequence of sets of labelled targets. However,
instead of computing the joint posterior distribution of the sequence
of sets of labelled targets, the usual approach in MTT is to build
trajectories based on the labeled multi-target filtering (or smoothing)
densities, i.e., the marginal distributions of these sets. 

Considering multi-target densities rather than the joint posterior
distribution, as required in a full Bayesian solution to MTT, has
the advantage of requiring lower computational resources. Yet, because
the joint posterior distribution of the trajectories is not contained
in multi-target densities, some problematic cases can arise.  For
instance, when new born targets are an independent and identically
distributed (IID) cluster RFS or Poisson RFS \cite{Mahler_book14,Williams15b},
labelled multitarget densities show total uncertainty in the associations
between labels and targets born at the same time, as will be explained
in Section \ref{subsec:Motivation-for-sets}. This label association
uncertainty implies that there can be a never ending track-switching
in the estimates for targets born at the same time unless some ad-hoc
mechanism is used. 

The above-mentioned problems of MTT based on labelled multitarget
densities can be solved by considering the joint density on the sequence
of sets of labelled targets at all time steps, as pointed out in \cite[Sec. II.B]{Vo14}
and used in \cite{Vu14}. This density is a valid representation of
the posterior distribution of the trajectories. Based on it, we can
answer all possible trajectory-related questions and optimally estimate
trajectories. However, as we will see, there is no need to artificially
identify targets through labelling, which increases the dimension
of the state, to estimate trajectories or answer trajectory-related
questions. Moreover, the arbitrariness of the labels prevents the
development of metrics with physical interpretation on the space of
sequences of labelled sets, as will be explained in Section \ref{subsec:Motivation-for-sets}.

In this paper, we propose the set of trajectories as the variable
of interest in MTT. In the standard dynamic model, targets are born,
move and die \cite{Mahler_book07} so a target trajectory is characterised
by a start time, a length and a sequence of target states. A set of
trajectories provides a minimal, unambiguous representation of the
MTT system at all time steps without arbitrary variables. This representation
enables us to define metrics with physical interpretation, such as
the ones in \cite{Rahmathullah16_prov2,Bento_draft16}, which are
important for evaluating algorithms and estimator design. Applying
Mahler's RFS framework \cite{Mahler_book14} to MTT with sets of trajectories,
we explain how to characterise a distribution over sets of trajectories
using a multitrajectory density, which corresponds to a multiobject
density in which the objects are trajectories. All information of
interest about the trajectories up to the current time is therefore
contained in the multitrajectory density given the available measurements.
Importantly, this multitrajectory density has significantly fewer
terms than a corresponding joint density over the sequence of labelled
sets, as will be analysed in Section \ref{subsec:Motivation-for-sets}.

One way of computing the required multitrajectory density is by using
the filtering equations with the set of trajectories as state variable.
The adoption of the set of trajectories as state variable constitutes
a natural and elegant analog of the RFS approach for multitarget filtering,
in which we aim to estimate the current set of targets, to multitarget
tracking. This representation therefore enables us to extend algorithms
for multitarget filtering, such as the probability hypothesis density
(PHD) filter \cite{Mahler03}, to multitarget tracking: the trajectory
PHD filter \cite{Angel18_c}.  Obviously, dealing with multitrajectory
densities is more challenging than dealing with multitarget densities.
Nevertheless, we think it is important to properly characterise the
full MTT problem in a Bayesian context. This is an important preliminary
step to develop approximations/algorithms that are suitable for answering
the trajectory-related questions that arise in different applications.
Therefore, the main purpose of this paper is to establish the theoretical
foundations to perform MTT using sets of trajectories, not the development
of efficient, practical algorithms.

In this paper, we also present the filtering equations and a conjugate
family of densities, in the spirit of \cite{Vo13}, for computing
the multitrajectory filtering density.  Finally, we also establish
the relation between the multitrajectory filtering density and the
multitarget filtering density.  Preliminary results on sets of trajectories
were provided in \cite{Svensson14}.  

The rest of the paper is organised as follows. In Section \ref{sec:General-concepts},
we define a set of trajectories and motivate its use as the state
variable. In Section \ref{sec:Calculation-posterior}, we provide
the recursive equations to calculate the multitrajectory filtering
 density. We analyse the relations among the proposed approach, labelled
approaches, the usual RFS tracking framework based on sets of targets
and classical MHT in Section \ref{sec:Relations-with-other}. Two
illustrative examples are provided in Section \ref{sec:Illustrative-example}
and concluding remarks are given in Section \ref{sec:Conclusions}.

\settowidth{\nomlabelwidth}{$\tau^{k}\left(\mathbf{X}\right)$}
\printnomenclature{}

\section{Sets of trajectories\label{sec:General-concepts}}

In this section, we introduce the variables, motivate why we propose
the set of trajectories as a state variable and indicate how to use
FISST for sets of trajectories. 

\subsection{State variables and notation\label{subsec:State-variables}}

A single target state $x\in D$\nomenclature{$x$}{Target state.},
where $D=\mathbb{R}^{n_{x}}$\nomenclature{$D$}{Target space.}, contains
the information of interest about the target, e.g., its position and
velocity. A set $\mathbf{x}$ of single target states belongs to $\mathcal{F}\left(D\right)$\nomenclature{$\mathbf{x}$}{Set of targets.}
where $\mathcal{F}\left(D\right)$ denotes the set of all finite subsets
of $D$. We are interested in representing the information on all
target trajectories, where a trajectory consists of a sequence of
target states that can start at any time step and end at any time
after it starts. Mathematically, a trajectory is represented as a
variable $X=\left(t,x^{1:i}\right)$\nomenclature{$X$}{$=\left(t,x^{1:i}\right)$. Trajectory state.}
where $t$ is the initial time step of the trajectory, $i$ is its
length and $x^{1:i}=\left(x^{1},...,x^{i}\right)$ denotes a sequence
of length $i$ that contains the target states at consecutive time
steps of the trajectory. 

We consider trajectories up to some finite time step\footnote{Considering a finite $k'$ ensures that the single trajectory space
$T_{\left(k'\right)}$ is locally compact, Hausdorff and second-countable.
These properties will be required to use finite set statistics, see
Section \ref{subsec:Probability-and-integration}. During filtering,
we are concerned with trajectories up the current time step, denoted
as $k$ in Section \ref{sec:Calculation-posterior}. Therefore, we
need $k'\geq k$ so that $T_{\left(k\right)}\subseteq T_{\left(k'\right)}$
and the single trajectory space contains the trajectories of interest.
We can achieve this by selecting $k\text{\textasciiacute}$ to be
orders of magnitude larger than the latest time that we will ever
consider in our applications, or by setting $k'=k$. Both choices
lead to the same filtering results. } $k'$. As a trajectory $\left(t,x^{1:i}\right)$ exists from time
step $t$ to $t+i-1$, variable $\left(t,i\right)$ belongs to the
set $I_{(k')}=\left\{ \left(t,i\right):0\leq t\leq k'\,\mathrm{and}\,1\leq i\leq k'-t+1\right\} $.
A single trajectory $X$ up to time step $k'$ therefore belongs to
the space $T_{\left(k'\right)}=\uplus_{\left(t,i\right)\in I_{(k')}}\left\{ t\right\} \times D^{i}$,
where $D^{i}$ represents\nomenclature{$T_{\left(k'\right)}$}{Trajectory space up to time step $k'$.}
$i$ Cartesian products of $D$ and $\uplus$ stands for disjoint
union\nomenclature{$\uplus$}{Disjoint union.}, which is used in this
paper to highlight that it is the union of disjoint sets. Similarly
to the set ${\bf x}$ of targets, we denote a set of trajectories
up to time step $k'$ as $\mathbf{X}\in\mathcal{F}\left(T_{\left(k'\right)}\right)$\nomenclature{$\mathbf{X}$}{Set of trajectories.}. 

Given a single target trajectory $X=\left(t,x^{1:i}\right)$, the
set $\tau^{k}\left(X\right)$ of the target state at time $k$ is
\[
\tau^{k}\left(X\right)=\begin{cases}
\left\{ x^{k+1-t}\right\}  & t\leq k\leq t+i-1\\
\emptyset & \mathrm{elsewhere}.
\end{cases}
\]
As the trajectory exists from time step $t$ to $t+i-1$, the set
is empty if $k$ is outside this interval. We employ the following
terminology: 
\begin{itemize}
\item A trajectory $X$ is present at time step $k$ if and only if $\left|\tau^{k}\left(X\right)\right|=1$. 
\item A surviving trajectory at time $k$ is a trajectory that is present
at times $k$ and $k-1$.
\end{itemize}
Given a set $\mathbf{X}$ of trajectories, the set $\tau^{k}\left(\mathbf{X}\right)$
\nomenclature{$\tau^{k}\left(\mathbf{X}\right)$}{Set of target states at time $k$ in $\mathbf{X}$.}of
target states at time $k$ is
\begin{equation}
\tau^{k}\left(\mathbf{X}\right)=\bigcup_{X\in\mathbf{X}}\tau^{k}\left(X\right).\label{eq:set_targets_formula}
\end{equation}
In RFS modelling, two or more targets at a given time cannot have
an identical state \cite[Sec 2.3]{Mahler_book14}. The corresponding
assumption for sets of trajectories is that any two trajectories $X,Y\in\mathbf{X}$
satisfy that $\tau^{k}\left(X\right)\cap\tau^{k}\left(Y\right)=\emptyset$
for all $k\in\mathbb{N}$. Note that this assumption ensures that
the cardinality $\left|\tau^{k}\left(\mathbf{X}\right)\right|$ of
$\tau^{k}\left(\mathbf{X}\right)$ represents the number of trajectories
present at time $k$. 
\begin{example}
\label{exa:Set_trajectories1}An example of a set of trajectories
with one-dimensional target states is $\mathbf{X}=\left\{ X_{1},X_{2},X_{3}\right\} $
with $X_{1}=\left(1,\left(1,1.5,2\right)\right)$, $X_{2}=\left(1,\left(0.5,0.625,0.75,0.875,1\right)\right)$
and $X_{3}=\left(2,\left(2.4,2.6,2.8,3\right)\right)$, which is illustrated
in Figure \ref{fig:Illustration_set_trajectories}(top). There is
one trajectory that starts at time 1 with length 3 and states $\left(1,1.5,2\right)$,
another that starts at time 1 with length 5 and states  $\left(0.5,0.625,0.75,0.875,1\right)$
and a third one that starts at time 2 with length 4 and states $\left(2.4,2.6,2.8,3\right)$.
We also have that, e.g., $\tau^{1}\left(\mathbf{X}\right)=\left\{ 1,0.5\right\} $
and $\tau^{5}\left(\mathbf{X}\right)=\left\{ 3,1\right\} $.

The information contained in $\mathbf{X}$ is also found in a sequence
of sets of labelled targets \cite{Vo14}, as illustrated in Figure
\ref{fig:Illustration_set_trajectories}(bottom). For example, squares,
crosses and circles can represent the target states with assigned
labels $l_{1}$, $l_{2}$ and $l_{3}$, respectively. However, the
assignment of labels to targets is arbitrary, as labels do not represent
any physical, meaningful quantity, so we can make any other association
(or assign completely different labels). For example, we can instead
consider that squares represent target states with label $l_{2}$,
crosses represent target states with label $l_{3}$ and circles represent
target states with label $l_{1}$ and we still represent the same
physical reality. In other words, with sets of trajectories, the mathematical
representation of the multiple trajectories is unique, but, with sequence
of sets of labelled targets, there are infinite representations, as
the labelling of the targets is arbitrary. The advantages of removing
these arbitrary labels are discussed in the next section. $\oblong$
\end{example}
\begin{center}
\begin{figure}
\begin{centering}
\includegraphics{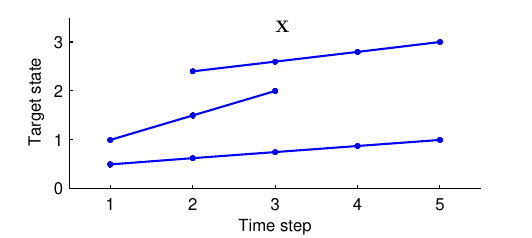}
\par\end{centering}
\begin{centering}
\includegraphics{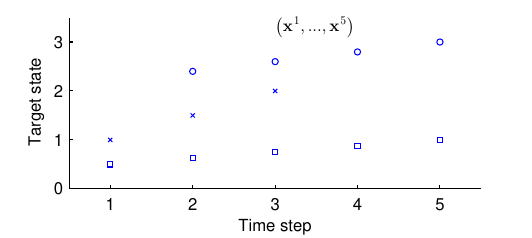}
\par\end{centering}
\caption{\label{fig:Illustration_set_trajectories}Illustration of set of trajectories
of Example \ref{exa:Set_trajectories1} (top) and an equivalent sequence
of sets of labelled targets where squares, crosses and circles represent
three different labels (bottom). }
\end{figure}
\par\end{center}

\subsection{Motivation for sets of trajectories\label{subsec:Motivation-for-sets}}

In this section, we motivate the importance of considering the multitrajectory
density, defined over sets of trajectories, in a full Bayesian approach
to MTT. 

In vector based state space models, it is often important to consider
the posterior density over the state trajectory, which contains the
states at all time steps, i.e., it is not sufficient to merely find
all the marginal densities of the states at all times \cite{Godsill04,Lindsten13}.
For example, the posterior density over the trajectory is necessary
to calculate the maximum a posterior (MAP) estimator of the trajectory
\cite{Miguez13} or to answer trajectory-related questions, e.g.,
what is the probability that the state was in a region at a time and
has moved to another region at a different time step? 

The same is true in MTT. Sometimes, it is not sufficient to calculate
multitarget densities at all time steps, we need a multitrajectory
density as it enables us to answer all trajectory related questions,
see Figure \ref{fig:llustration-trajectory-questions}. For example,
what is the probability that a target was in Madrid four hours ago
and is currently in Gothenburg? This problem of defining a multitrajectory
density to answer trajectory related questions has received little
attention in the MTT literature and, in this section, we present arguments
that support the idea that such a multitrajectory density should be
defined on the space of sets of trajectories. To this end, we proceed
to review some characteristics of labelled RFS approaches to MTT and
indicate their shortcomings. 

In the typical labelled RFS approach, we consider the sequence of
multitarget densities on the set of labelled targets up to the current
time step, which can be used to estimate suitable trajectories in
many situations. However, this sequence of  multitarget densities
does not contain all available information and does not let us answer
trajectory related questions, which are of key importance in tracking.
As we illustrate next, a lack of complete information is particularly
problematic when there is an unknown association between (unlabelled)
targets states and labels. Two examples when this happens is when
the birth process is an IID cluster RFS, in which the new born targets
are IID given the cardinality \cite[Sec. 4.3.2]{Mahler_book14}, and
when targets get in close proximity and then separate \cite{Angel14,Blom08,Blom09}.
This is an important weakness since the Poisson RFS, a specific type
of IID cluster RFS, is a commonly used birth model \cite[Sec. 14.2.1]{Mahler_book07}\cite{Williams15b}.
We illustrate the problem of the IID cluster birth RFS in a simple
example. 

\begin{figure}
\begin{centering}
\subfloat[]{\begin{centering}
\includegraphics[scale=0.29]{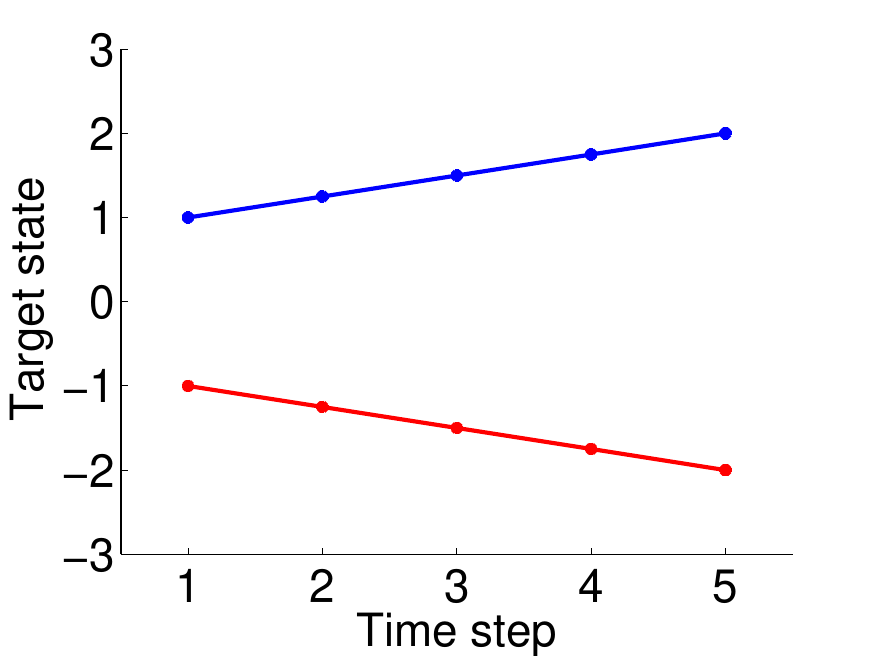}
\par\end{centering}
}\subfloat[]{\begin{centering}
\includegraphics[scale=0.29]{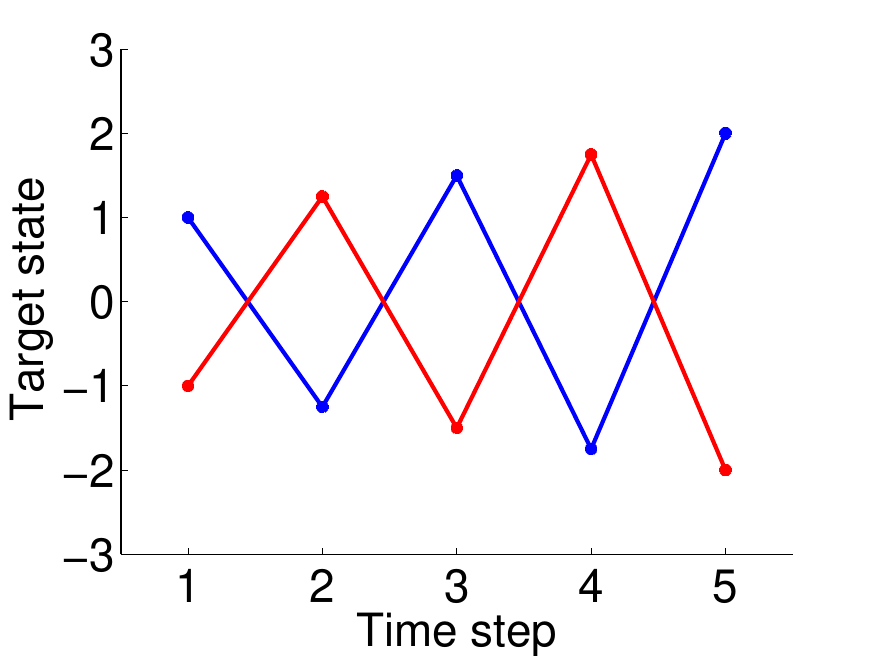}
\par\end{centering}
}
\par\end{centering}
\caption{\label{fig:Equally-likely-trajectories}Illustration of two possible
trajectory states from the sequence of labelled multitarget filtering/smoothing
densities of Example \ref{exa:Label_mixing}. If we use the multitrajectory
filtering density (defined on the space of sets of trajectories),
the only possible state is the one in the left figure. }
\end{figure}

\begin{example}
\label{exa:Label_mixing}Let us consider the following scenario. New
born targets at time 1 are modelled by a labelled multi-Bernoulli
RFS with two components. According to the labelling convention in
\cite[Sec. IV.D]{Vo13}, the labels of the two components are $a=\left(1,1\right)$
and $b=\left(1,2\right)$, where the first component of the label
is the time of birth and the second, a unique index to distinguish
targets born at the same time. Both components have existence probability
one and the same Gaussian density with a certain mean and variance.
Note that if we remove the labels, the new born targets are an IID
cluster RFS. Targets move independently with a given transition density
and probability of survival one and no more targets can be born afterwards.
Target states (without labels) are observed directly using the standard
measurement model with no clutter, probability of detection 1 and
negligible noise \cite{Mahler_book14}. This implies that, at each
time $k\in\left\{ 1,...,5\right\} $,  the measurement set is $\left\{ z_{1}^{k},z_{2}^{k}\right\} =\left\{ x_{1}^{k},x_{2}^{k}\right\} $,
where $\left\{ x_{1}^{k},x_{2}^{k}\right\} $ is the set of unlabelled
targets at time step $k$. We also assume that the single target transition
density is such that target movements from $z_{1}^{k-1}$ to $z_{2}^{k}$
and from $z_{2}^{k-1}$ to $z_{1}^{k}$ do not occur. We have described
a toy example without uncertainties in the two trajectories, but,
as we will see, it is still challenging to handle using labelled multitarget
densities. 

The  labelled multitarget filtering density $\pi^{k}\left(\cdot\right)$
at time $k$, which is given by the $\delta$-generalised labelled
multi-Bernoulli ($\delta$-GLMB) filter \cite{Vo13} and coincides
with the smoothing solution, is
\begin{align}
 & \pi^{k}\left(\left\{ \left(x_{1}^{k},l_{1}^{k}\right),\left(x_{2}^{k},l_{2}^{k}\right)\right\} \right)\nonumber \\
 & =\frac{1}{2}\left(\delta_{z_{1}^{k}}\left(x_{1}^{k}\right)\delta_{z_{2}^{k}}\left(x_{2}^{k}\right)+\delta_{z_{1}^{k}}\left(x_{2}^{k}\right)\delta_{z_{2}^{k}}\left(x_{1}^{k}\right)\right)\nonumber \\
 & \quad\times\left(\delta_{a}\left[l_{1}^{k}\right]\delta_{b}\left[l_{2}^{k}\right]+\delta_{a}\left[l_{2}^{k}\right]\delta_{b}\left[l_{1}^{k}\right]\right)\label{eq:PDF_example_label_mixing}
\end{align}
and zero for other sets of labelled targets. Notation $\delta_{y}\left(\cdot\right)$
and $\delta_{y}\left[\cdot\right]$ represent the Dirac and Kronecker
delta centered at $y$, respectively, and $\left(x_{j}^{k},l_{j}^{k}\right)$
represents the state and label of target $j$ at time $k$. Evidently,
the multitarget filtering/smoothing density at any time $k$ contains
the information that there are targets located at $\left\{ z_{1}^{k},z_{2}^{k}\right\} $,
but labelling them as $\left\{ \left(z_{1}^{k},a\right),\left(z_{2}^{k},b\right)\right\} $
is as likely as $\left\{ \left(z_{1}^{k},b\right),\left(z_{2}^{k},a\right)\right\} $.
This result holds even though the transition density indicates that
movements from $z_{1}^{k-1}$ to $z_{2}^{k}$ and from $z_{2}^{k-1}$
to $z_{1}^{k}$ do not happen. Therefore, if we follow the usual procedure
and build possible trajectories by linking target states and labels
by using the filtering/smoothing multitarget densities in isolation,
we obtain many possible sequences $\left(\mathbf{x}^{1},...,\mathbf{x}^{5}\right)$
of labelled sets, which represent trajectories. For example, as illustrated
in Figure \ref{fig:Equally-likely-trajectories}, we can have $\mathbf{x}^{k}=\left\{ \left(a,z_{1}^{k}\right),\left(b,z_{2}^{k}\right)\right\} $
for all $k$, see Figure \ref{fig:Equally-likely-trajectories}(a),
but also $\mathbf{x}^{k}=\left\{ \left(a,z_{1}^{k}\right),\left(b,z_{2}^{k}\right)\right\} $
for odd $k$ and $\mathbf{x}^{k}=\left\{ \left(b,z_{1}^{k}\right),\left(a,z_{2}^{k}\right)\right\} $
for even $k$, see Figure \ref{fig:Equally-likely-trajectories}(b).
Consequently, we cannot tell how targets move between different time
steps. 

We recall that the total ambiguity in label-to-target associations
in this example happens due to the type of birth model and the use
of multitarget filtering/smoothing densities. Therefore, in this case,
estimators based on these densities do not have enough information
to link the target states and form suitable trajectories. In practice,
one can employ  pragmatic fixes, which can also be used with unlabelled
filters \cite{Williams15b}, to estimate sensible trajectories in
this example. For instance, one can use the dynamic model or the metadata
associated to the filters, such as the history of data associations.
Nevertheless, a full Bayesian methodology to MTT should not rely on
pragmatic fixes, but on densities that contain the required information.
 $\oblong$
\end{example}

The previous issues of performing MTT using labelled multitarget densities
can be solved by considering the joint density over the sequence of
sets of labelled targets \cite{Vu14}. This density contains full
trajectory information so it enables us to answer all trajectory-related
questions. However, this representation has two drawbacks. The first
one is that, due to the inclusion of arbitrary labels, this sequence
of sets of labelled targets does not uniquely represent the underlying
physical reality, see Example \ref{exa:Set_trajectories1}. This implies
that we cannot define metrics with physical interpretation, on the
space of sequences of sets of labelled targets, because, due to the
identity property of metrics\footnote{The identity property says that a metric $d\left(\cdot,\cdot\right)$
on a certain space must satisfy $d\left(x,y\right)=0$ if and only
if $x=y$ for any two elements $x$ and $y$ in the space. } \cite{Schuhmacher08}, we always obtain a non-zero distance/error
between two different sequences of sets of labelled targets describing
the same trajectories. For instance, changing the crosses and circles
in Figure \ref{fig:Illustration_set_trajectories}(bottom), we represent
the same trajectories but the distance between this sequence of labelled
sets and the (equivalent) original one is non-zero for any metric
on the space of sequences of sets of labelled targets. This means
that evaluating the performance of MTT algorithms using a metric on
the space of sequences of sets of labelled targets is not useful,
as it can provide a non-zero distance/error when there is no estimation
error. The second drawback is that the explicit expressions of the
joint densities over the sequence of labelled targets are cumbersome,
as is illustrated in the next example.
\begin{example}
\label{exa:Label_mixing_joint_PDF}The joint density over the sequence
of labelled sets for the first two time steps in the scenario described
in Example \ref{exa:Label_mixing} is explicitly written as \cite{Vo14}
\begin{align*}
 & \pi^{1:2}\left(\left\{ \left(x_{1}^{1},l_{1}^{1}\right),\left(x_{2}^{1},l_{2}^{1}\right)\right\} ,\left\{ \left(x_{1}^{2},l_{1}^{2}\right),\left(x_{2}^{2},l_{2}^{2}\right)\right\} \right)\\
 & =\frac{1}{2}\left[\left(\delta_{\left(a,b,a,b\right)}\left[l_{1}^{1},l_{2}^{1},l_{1}^{2},l_{2}^{2}\right]+\delta_{\left(a,b,a,b\right)}\left[l_{2}^{1},l_{1}^{1},l_{2}^{2},l_{1}^{2}\right]\right)\right.\\
 & \times\left(\delta_{\left(z_{1}^{1},z_{2}^{1},z_{1}^{2},z_{2}^{2}\right)}\left(x_{1:2}^{1:2}\right)+\delta_{\left(z_{2}^{1},z_{1}^{1},z_{2}^{2},z_{1}^{2}\right)}\left(x_{1:2}^{1:2}\right)\right)\\
 & +\left(\delta_{\left(a,b,a,b\right)}\left[l_{1}^{1},l_{2}^{1},l_{2}^{2},l_{1}^{2}\right]+\delta_{\left(a,b,a,b\right)}\left[l_{2}^{1},l_{1}^{1},l_{1}^{2},l_{2}^{2}\right]\right)\\
 & \left.\times\left(\delta_{\left(z_{2}^{1},z_{1}^{1},z_{1}^{2},z_{2}^{2}\right)}\left(x_{1:2}^{1:2}\right)+\delta_{\left(z_{1}^{1},z_{2}^{1},z_{2}^{2},z_{1}^{2}\right)}\left(x_{1:2}^{1:2}\right)\right)\right]
\end{align*}
where $x_{1:2}^{1:2}$ is used in this example to denote $\left(x_{1}^{1},x_{2}^{1},x_{1}^{2},x_{2}^{2}\right)$
and $\pi^{1:2}\left(\cdot\right)$ is zero for other sequences of
labelled sets. As required, according to this density, $z_{1}^{1}$
and $z_{2}^{1}$ can only be linked with $z_{1}^{2}$ and $z_{2}^{2}$,
respectively. However, these links arise in multiple combinations
of the sequence of labelled sets $\left(\left\{ \left(x_{1}^{1},l_{1}^{1}\right),\left(x_{2}^{1},l_{2}^{1}\right)\right\} ,\left\{ \left(x_{1}^{2},l_{1}^{2}\right),\left(x_{2}^{2},l_{2}^{2}\right)\right\} \right)$.
In fact, for a sequence of length $k$, in this example we have that
the number of terms in the joint density is $2^{k+1}$, which represents
the number of possible associations of target states to measurements
($2^{k}$) and 2 possible ways of labelling them. Even though we only
consider two time steps, the above expression already contains 8 terms
and the corresponding expression for a sequence of length five contains
64 terms. Due to this exponential increase in the number of terms,
the inclusion of labels makes the explicit expression cumbersome even
for relatively short sequences. $\oblong$
\end{example}
The mentioned drawbacks of sequences of labelled sets can be solved
by using sets of trajectories. Sets of trajectories do not include
arbitrary parameters so we can develop metrics, such as the ones proposed
in \cite{Rahmathullah16_prov2,Bento_draft16}. Though these metrics
are not metrics on the space of sequences of sets of labelled targets,
they can be used to evaluate MTT algorithms based on sequences of
sets of labelled targets, by representing the resulting estimates
in terms of sets of (unlabelled) trajectories. In addition, a (multiobject)
density on the set of trajectories enables us to answer all possible
trajectory related questions with a more compact representation, as
illustrated in the next example.
\begin{example}
As we will explain in this paper, by applying Mahler's RFS framework
to set of trajectories, the multitrajectory density for the five time
steps in Example \ref{exa:Label_mixing} is 
\begin{align*}
 & \pi^{5}\left(\left\{ \left(t_{1},x_{1}^{1:i_{1}}\right),\left(t_{2},x_{2}^{1:i_{2}}\right)\right\} \right)\\
 & =\left(\delta_{z_{1}^{1:5}}\left(x_{1}^{1:i_{1}}\right)\delta_{z_{2}^{1:5}}\left(x_{2}^{1:i_{1}}\right)+\delta_{z_{1}^{1:5}}\left(x_{2}^{1:i_{1}}\right)\delta_{z_{2}^{1:5}}\left(x_{1}^{1:i_{1}}\right)\right)\\
 & \quad\times\delta_{1}\left[t_{1}\right]\delta_{1}\left[t_{2}\right]\delta_{5}\left[i_{1}\right]\delta_{5}\left[i_{2}\right],
\end{align*}
and zero for other sets of trajectories. This multitrajectory density
has complete trajectory information with a significant decrease in
the number of terms compared to the joint density over the sequence
of labelled sets, 2 terms versus 64, as pointed out in Example \ref{exa:Label_mixing_joint_PDF}.
$\oblong$
\end{example}

Once the full Bayesian problem is properly characterised, we can develop
algorithms/approximations to handle the trajectory-related questions
of the application at hand. For example, if our application only requires
us to estimate the number of targets and their positions at the current
time, it is enough to consider the filtering multitarget density at
the current time, as in the usual RFS approach.

\subsection{Probability and integration\label{subsec:Probability-and-integration}}

In this paper, probability and integration are defined using finite
set statistics (FISST) \cite{Mahler_book07,Mahler_book14}, which
is related to measure theory \cite{Vo05}. Even though FISST usually
considers sets of targets, it can be applied to sets of trajectories
by changing the single object state (targets by trajectories) and
single object integrals (single target integrals by single trajectory
integrals). In Appendix \ref{sec:Appendix_FISST}, we explain why
we can use FISST with sets of trajectories and how to obtain the corresponding
single-trajectory integrals and set integrals, which are given in
the following. 

Given a real-valued function $\pi\left(\cdot\right)$ on the single
trajectory space $T_{\left(k'\right)}$, its integral is 
\begin{align}
\int\pi\left(X\right)dX & =\sum_{\left(t,i\right)\in I_{(k')}}\int\pi\left(t,x^{1:i}\right)dx^{1:i}.\label{eq:single_trajectory_integral}
\end{align}
This integral goes through all possible start times, lengths and target
states of the trajectory. Given a real-valued function $\pi\left(\cdot\right)$
on the space $\mathcal{F}\left(T_{\left(k'\right)}\right)$ of sets
of trajectories, its set integral is \cite{Mahler_book07}: 
\begin{align}
\int\pi\left(\mathbf{X}\right)\delta\mathbf{X} & =\sum_{n=0}^{\infty}\frac{1}{n!}\int\pi\left(\left\{ X_{1},...,X_{n}\right\} \right)dX_{1:n}\label{eq:set_integral_trajectory}
\end{align}
where $X_{1:n}=\left(X_{1},...,X_{n}\right)$. Note that, if $\pi\left(\cdot\right)$
is a multitrajectory density, then, $\pi\left(\cdot\right)\geq0$
and its set integral is one.  

We can also use set integrals to calculate the probability that an
RFS of trajectories belongs to a certain region. In order to do so,
we define a mapping $\chi:\uplus_{n\text{=0}}^{\infty}T_{\left(k'\right)}^{n}\rightarrow\mathcal{F}\left(T_{\left(k'\right)}\right)$\nomenclature{$\chi\left(\cdot\right)$}{Mapping from vectors to sets.}
of sequences of trajectories to sets of trajectories such that $\chi\left(\left(X_{1},...,X_{n}\right)\right)=\left\{ X_{1},...,X_{n}\right\} $.
Given a region $A=\uplus_{n\text{=0}}^{\infty}A_{n}$ where $A_{n}\subseteq\mathcal{F}\left(T_{\left(k'\right)}\right)$
is a set that contains sets with $n$ elements in $T_{\left(k'\right)}$,
the probability that $\mathbf{X}$ belongs to $A$ is 
\begin{align}
P\left(\mathbf{X}\in A\right) & =\sum_{n=0}^{\infty}\frac{1}{n!}\int_{\chi^{-1}\left(A_{n}\right)}\pi\left(\left\{ X_{1},...,X_{n}\right\} \right)dX_{1:n}.\label{eq:probability_evaluation_trajectories}
\end{align}
where $\pi\left(\cdot\right)$ is the multitrajectory density of $\mathbf{X}$.
Equation (\ref{eq:probability_evaluation_trajectories}) is proved
in Appendix \ref{sec:Appendix_Integration_janossy}. For instance,
if $A_{2}=\left\{ \left\{ X_{1},X_{2}\right\} :X_{1}\in B_{1},X_{2}\in B_{2}\right\} $
where $B_{1}\subseteq T_{\left(k'\right)}$ and $B_{2}\subseteq T_{\left(k'\right)}$,
then $\chi^{-1}\left(A_{2}\right)=\left\{ \left(X_{1},X_{2}\right):X_{1}\in B_{1},X_{2}\in B_{2}\,\mathrm{or}\,X_{2}\in B_{1},X_{1}\in B_{2}\right\} $.
Calculating (\ref{eq:probability_evaluation_trajectories}) for $A_{2}$,
we obtain the probability that there are two trajectories, one in
region $B_{1}$ and another one in region $B_{2}$. Equation (\ref{eq:probability_evaluation_trajectories})
is necessary to obtain certain probabilities of interest, for example,
the probability that there is a number of trajectories present at
a certain time instant or a number of targets in a given region at
a given time. The cardinality distribution $\rho\left(\cdot\right)$
indicates the probability that $n$ trajectories have existed at all
times 
\begin{align}
\rho\left(n\right) & =P\left(\mathbf{X}\in\chi\left(T_{\left(k'\right)}^{n}\right)\right)\nonumber \\
 & =\frac{1}{n!}\int\pi\left(\left\{ X_{1},...,X_{n}\right\} \right)dX_{1:n},\label{eq:cardinality_distribution_trajectories}
\end{align}
which is analogous for RFSs of targets \cite[Eq. (11.115)]{Mahler_book07}. 
\begin{example}
We consider a multitrajectory density $\pi\left(\cdot\right)$ such
that
\begin{align*}
\pi\left(\left\{ \left(1,x_{1}^{1:2}\right)\right\} \right) & =0.9\mathcal{N}\left(x_{1}^{1:2};\left(10,11\right),\left[\begin{array}{cc}
1 & 1\\
1 & 2
\end{array}\right]\right)\\
\pi\left(\left\{ \left(1,x_{1}^{1:2}\right),\left(2,x_{2}^{1}\right)\right\} \right) & =0.1\mathcal{N}\left(x_{1}^{1:2};\left(10,11\right),\left[\begin{array}{cc}
1 & 1\\
1 & 2
\end{array}\right]\right)\\
 & \quad\times\mathcal{N}\left(x_{2}^{1};100,1\right),
\end{align*}
where $\mathcal{N}\left(\cdot;\overline{x},P\right)$ denotes a Gaussian
density with mean $\overline{x}$ and covariance matrix $P$, and
$\pi\left(\cdot\right)$ is zero for other sets of trajectories. From
(\ref{eq:cardinality_distribution_trajectories}), we see that the
probability that there is one trajectory is 0.9 and the probability
for two trajectories is 0.1. The probability that there is only one
trajectory and this trajectory starts at time step 1 in a region $B_{1}\subseteq D$
and moves to a region $B_{2}\subseteq D$ at the next time step can
be obtained by integrating $\pi\left(\cdot\right)$ over the region
$A_{1}=\chi\left(\left\{ 1\right\} \times B\times\uplus_{i=0}^{k'-2}D^{i}\right)$,
where $B=B_{1}\times B_{2}$. That is, region $A_{1}$ considers start
time 1 with the first two states belonging to $B$. Then, the trajectory
can die at any moment afterwards and, when it is present, its state
at a particular time belongs to $D$. Then, 
\begin{align}
P\left(\mathbf{X}\in A_{1}\right) & =0.9\int_{B}\mathcal{N}\left(x_{1}^{1:2};\left(10,11\right),\left[\begin{array}{cc}
1 & 1\\
1 & 2
\end{array}\right]\right)dx_{1}^{1:2}\label{eq:probability_region_example}
\end{align}
where we have used (\ref{eq:probability_evaluation_trajectories})
and that $\chi^{-1}\left(A_{1}\right)=\left\{ 1\right\} \times B\times\uplus_{i=0}^{k'-2}D^{i}$.
Note that, using the notation in (\ref{eq:probability_evaluation_trajectories}),
$A=A_{1}$, as the integration region in this example only contains
sets of cardinality $1$ and, therefore, (\ref{eq:probability_region_example})
only considers the term that corresponds to $n=1$ in (\ref{eq:probability_evaluation_trajectories}).
$\oblong$ 
\end{example}

\section{Filtering recursion for RFSs of trajectories\label{sec:Calculation-posterior}}

In this section, we present the filtering recursion for RFS of trajectories.
We first present the dynamic model of the trajectories in Section
\ref{subsec:Dynamic-model}. Then, for this dynamic model, we present
the filtering recursion for a general measurement model and for the
standard measurement model in Sections \ref{subsec:General-equations}
and \ref{subsec:Conjugate-form}, respectively. We discuss some practical
considerations in Section \ref{subsec:Practical-considerations}. 

\subsection{Dynamic model\label{subsec:Dynamic-model}}

\begin{figure}
\begin{centering}
\includegraphics[scale=0.55]{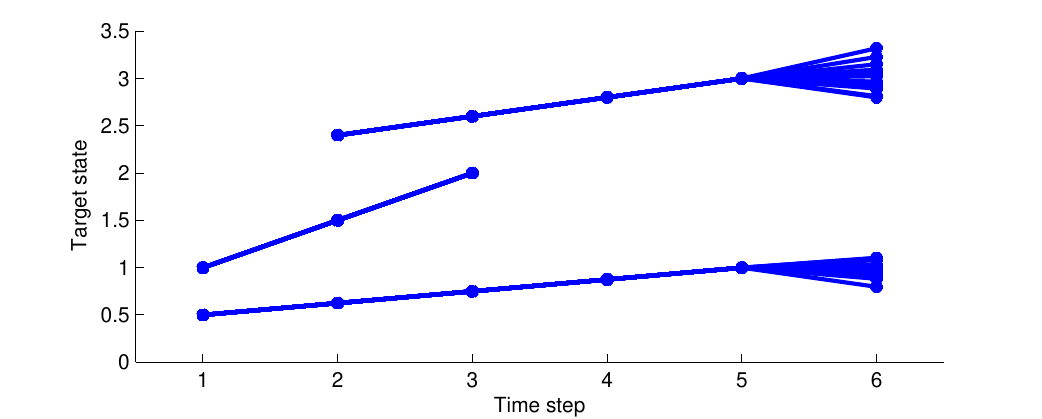}
\par\end{centering}
\caption{\label{fig:-20-posssible-realisations} Dynamic model: 20 possible
realisations of the random set of trajectories at time 6 given the
set of trajectories of Figure \ref{fig:Illustration_set_trajectories}
at time 5. No new born targets are considered and $p_{S}\left(\cdot\right)=1$.}
\end{figure}

We consider the conventional assumptions for the dynamic model used
in the RFS framework \cite{Mahler_book07}:
\begin{itemize}
\item Given the current multitarget state ${\bf x}$, each target $x\in{\bf x}$
survives with probability $p_{S}\left(x\right)$ and moves to a new
state with a transition density $g\left(\cdot\left|x\right.\right)$,
or dies with probability $1-p_{S}\left(x\right)$.
\item The multitarget state at the next time step is the union of the surviving
targets and new targets, which are born independently of the rest
with a multitarget density $\beta_{\tau}\left(\cdot\right)$.
\end{itemize}
In this paper, we use the subindex $\tau$ in multitarget densities
to differentiate them from multitrajectory densities. The previous
parameters of the dynamic model can change with time but we omit time
dependence for notational simplicity. Note that, as in filtering RFS
of targets, this model implies that the number of trajectories and
new born targets at each time step is unknown. As we will see, this
dynamic model gives rise to a transition multitrajectory density $f^{k}\left(\cdot\left|\cdot\right.\right)$
for the set of trajectories at time $k$, which includes all trajectories
that have ever been present.
\begin{example}
We proceed to illustrate how the set of trajectories of Example \ref{exa:Set_trajectories1},
which is represented in Figure \ref{fig:Illustration_set_trajectories},
evolves with time. We consider that the current time step is $k=5$
and discuss what the set may look like at time $k=6$. Trajectory
$X_{1}$, which is not present at time 5, remains unaltered. Trajectory
$X_{2}$ survives with probability $p_{S}\left(1\right)$, which means
that it becomes $X_{2}=\left(1,\left(0.5,0.625,0.75,0.875,1,y\right)\right)$
with $y\sim g\left(\cdot\left|1\right.\right)$, or remains unaltered
with probability $1-p_{S}\left(1\right)$. An analogous behaviour
is shown by $X_{3}$. The new set is guaranteed to contain these three
trajectories plus new trajectories determined by the new born targets,
generated from the birth process $\beta_{\tau}\left(\cdot\right)$,
and the time of appearance $6$. For illustration, in Figure \ref{fig:-20-posssible-realisations},
we show 20 realisations of the random set of trajectories at time
6 using $g\left(y\left|x\right.\right)=\mathcal{N}\left(y;x,0.01\right)$,
probability of survival one at time 5 ($p_{S}\left(\cdot\right)=1$),
and no new born targets at time 6. $\oblong$ 
\end{example}
For this dynamic model, we present the filtering recursion for a general
measurement model in Section \ref{subsec:General-equations} and for
the standard measurement model in Section \ref{subsec:Conjugate-form}.

\subsection{Filtering with a general measurement model\label{subsec:General-equations}}

The set of targets at time $k$ is observed through noisy measurements
giving rise to the likelihood $\ell^{k}\left(\cdot\right)$, where
we omit the value of the measurement for notational simplicity \cite{Angel16}.
By using FISST \cite{Mahler_book14}, the multitrajectory filtering
density $\pi^{k}\left(\cdot\right)$ at time $k$, i.e., the multitrajectory
density of the set of trajectories up to time step $k$ given the
sequence of measurements up to time step $k$, can be calculated recursively
using the prediction and update equations,
\begin{align}
\pi^{k|k-1}\left(\mathbf{X}\right) & =\int f^{k}\left(\mathbf{X}\left|\mathbf{Y}\right.\right)\pi^{k-1}\left(\mathbf{Y}\right)\delta\mathbf{Y}\label{eq:prediction_trajectories}\\
\pi^{k}\left(\mathbf{X}\right) & \propto\ell^{k}\left(\mathbf{X}\right)\pi^{k|k-1}\left(\mathbf{X}\right).\label{eq:update_trajectories}
\end{align}
Here, $\propto$ means ``is proportional to'' and $\pi^{k|k-1}\left(\cdot\right)$
is referred to as the predicted multitrajectory density at time $k$,
which represents the density of the set of trajectories at time $k$
given the sequence of measurements up to time $k-1$.  It should
be noted that $\mathbf{X}$ drawn from $\pi^{k}\left(\cdot\right)$
or $\pi^{k|k-1}\left(\cdot\right)$ is a set of trajectories in the
time interval 1 to $k$. That is, we have that $\pi^{k}\left(\mathbf{X}\right)=0$
and $\pi^{k|k-1}\left(\mathbf{X}\right)=0$ if $\mathbf{X}$ contains
a trajectory that is present at a time step that is higher than $k$.
After time $k$, some trajectories in $\mathbf{X}$ may be extended,
as illustrated in Figure \ref{fig:-20-posssible-realisations}, and
new trajectories may appear. These properties imply that we only need
to compute the set integrals up to time step $k$. In theory, set
integrals are defined up to some finite time step $k\lyxmathsym{\textquoteright}$,
see (\ref{eq:single_trajectory_integral}), but, as long as $k\lyxmathsym{\textquoteright}>k$,
the actual value of $k\lyxmathsym{\textquoteright}$ is irrelevant
and does not have to be specified. 

For general track-before-detect measurement models, the likelihood
$\ell^{k}\left(\cdot\right)$ cannot be simplified so we just write
the update as in (\ref{eq:update_trajectories}) \cite{Angel16}.
The following theorem indicates how to evaluate the prediction (\ref{eq:prediction_trajectories})
more explicitly. 
\begin{thm}
\label{thm:General_prediction}We consider the conventional dynamic
model, which includes the functions $p_{S}\left(\cdot\right)$, $g\left(\cdot\left|\cdot\right.\right)$
and $\beta_{\tau}\left(\cdot\right)$, explained at the beginning
of Section \ref{subsec:Dynamic-model}, and trajectories up to time
step $k$. Then, given a set $\mathbf{W}$ of new born trajectories
at time $k$, a set $\mathbf{X}$ of trajectories present at times
$k-1$ and $k$, a set $\mathbf{Y}$ of trajectories present at time
$k-1$ but not present at time $k$ and a set $\mathbf{Z}$ of trajectories
present at a time before $k-1$ but not at $k-1$, the predicted multitrajectory
density $\pi^{k|k-1}\left(\cdot\right)$ at time $k$ is 
\begin{align*}
 & \pi^{k|k-1}\left(\mathbf{W}\uplus\mathbf{X}\uplus\mathbf{Y}\uplus\mathbf{Z}\right)\\
 & =\prod_{\left(t,x^{1:i}\right)\in\mathbf{X}}\left(g\left(x^{i}\left|x^{i-1}\right.\right)p_{S}\left(x^{i-1}\right)\right)\prod_{\left(t,x^{1:i}\right)\in\mathbf{Y}}\left(1-p_{S}\left(x^{i}\right)\right)\\
 & \quad\times\pi^{k-1}\left(\cup_{\left(t,x^{1:i}\right)\in\mathbf{X}}\left\{ \left(t,x^{1:i-1}\right)\right\} \uplus\mathbf{Y}\uplus\mathbf{Z}\right)\beta_{\tau}\left(\tau^{k}\left(\mathbf{W}\right)\right).
\end{align*}
\end{thm}
Theorem \ref{thm:General_prediction} is proved in Appendix \ref{sec:Appendix_general_prediction}.
We first clarify that if $\left(t,x^{1:i}\right)\in\mathbf{W}$, then,
$t=k$, $i=1$; if it belongs to $\mathbf{X}$, then $t<k$, $i=k-t+1$;
if it belongs to $\mathbf{Y}$, then $t<k$, $i=k-t$; and finally,
if it belongs to $\mathbf{Z}$, then, $t<k-1$, $i<k-t$. To evaluate
the predicted multitrajectory density at time $k$, we multiply the
following terms: multitrajectory filtering density $\pi^{k-1}\left(\cdot\right)$
for trajectories present at previous times, $g\left(\cdot|\cdot\right)$
and $p_{S}\left(\cdot\right)$ for surviving trajectories, $\left(1-p_{S}\left(\cdot\right)\right)$
for trajectories present at time $k-1$ but not present at $k$ and
the multitarget density $\beta_{\tau}\left(\cdot\right)$ for new
born targets. 

\subsection{Filtering with the standard measurement model\label{subsec:Conjugate-form}}

In this section, we present the following key result: for the standard
(point target) measurement model (\ref{eq:likelihood_radar}) and
the birth model (\ref{eq:birth_model_radar}), the multitrajectory
filtering and predicted densities at all time steps have the same
form, which is a multi-Bernoulli mixture in which the existence probabilities
are either 0 or 1, which we refer to as $\mathrm{MBM}_{01}$ \cite[Sec. IV]{Angel18_b}.
Therefore, the $\mathrm{MBM}_{01}$ multitrajectory density is conjugate
with respect to the standard measurement likelihood \cite{Vo13}.
How to obtain these multitrajectory densities recursively is indicated
by Lemmas \ref{lem:Prediction_radar} and \ref{lem:Update_radar},
which give rise to the trajectory $\mathrm{MBM}_{01}$ filter. We
use the multiobject exponential notation $h^{\mathbf{x}}=\prod_{x\in\mathbf{x}}h\left(x\right)$
\nomenclature{$h^{\mathbf{x}}$}{Multiobject exponential for function $h$ and set $\mathbf{x}$.}where
$h$ is a real-valued function and $h^{\emptyset}=1$ by convention
\cite{Vo13}. 

The standard measurement model \cite{Mahler_book14} is described
as:
\begin{itemize}
\item For a given multi-target state $\mathbf{x}$ at time $k$, each target
state $x\in\mathbf{x}$ is either detected with probability $p_{D}\left(x\right)$
and generates one measurement with density $l\left(\cdot|x\right)$,
or missed with probability $1-p_{D}\left(x\right)$. 
\item The measurement set $\mathbf{z}^{k}$ is the union of the target-generated
measurements and Poisson clutter with intensity function $\kappa\left(\cdot\right)$. 
\end{itemize}
We proceed to write the resulting likelihood \cite[Eq. (7.21)]{Mahler_book14}
in terms of sets of trajectories in a form that will be useful in
the rest of the section.

Given a set $\mathbf{X}\uplus\mathbf{Y}$ of trajectories such that
$\mathbf{X}=\left\{ X_{1},...,X_{n}\right\} $ with $n$ present trajectories
at time $k$ and a set $\mathbf{Y}$ of trajectories with no present
trajectories at time $k$, the measurement set $\mathbf{z}^{k}=\left\{ z_{1}^{k},...,z_{m}^{k}\right\} $
only depends on $\tau^{k}\left(\mathbf{X}\right)$ as explained before.
Then, the likelihood at time $k$ for the standard measurement model
is
\begin{align}
\ell^{k}\left(\mathbf{X}\uplus\mathbf{Y}\right) & =e^{-\int\kappa\left(z\right)dz}\kappa^{\mathbf{z}^{k}}\sum_{\theta\in\Theta_{n,m}}\prod_{j=1}^{n}\psi_{\mathbf{z}^{k}}\left(X_{j}|\theta_{j}\right),\label{eq:likelihood_radar}
\end{align}
where $\Theta_{n,m}$ denotes all data association hypotheses for
$n$ targets (which correspond to $n$ present trajectories) and $m$
measurements. More specifically, for $\theta=\left(\theta_{1},...,\theta_{n}\right)\in\Theta_{n,m}$,
$\theta_{i}=j$ if the $i$th present target is associated with the
$j$th measurement or $0$ if it is undetected. Due to the properties
of the standard measurement model, we have that, if $\theta_{i}=\theta_{j}>0$,
then $i=j$. Also,
\begin{align}
\psi_{\mathbf{z}^{k}}\left(t,x^{1:i}|\theta_{j}\right) & =\left\{ \begin{array}{cc}
\frac{p_{D}\left(x^{k-t+1}\right)l\left(z_{\theta_{j}}^{k}|x^{k-t+1}\right)}{\kappa\left(z_{\theta_{j}}^{k}\right)} & \theta_{j}>0\\
1-p_{D}\left(x^{k-t+1}\right) & \theta_{j}=0,
\end{array}\right.\label{eq:Psi_radar}
\end{align}
where $x^{k-t+1}$ is the state of trajectory $\left(t,x^{1:i}\right)$
that corresponds to time step $k$. If $\mathbf{X}=\emptyset$, then
$\ell^{k}\left(\mathbf{Y}\right)=e^{-\int\kappa\left(z\right)dz}\kappa^{\mathbf{z}^{k}}$. 

In order to obtain the explicit recursion, we assume that the targets
can be born from $b$ densities $\beta_{1}\left(\cdot\right)$, ...,
$\beta_{b}\left(\cdot\right)$ with a weight $w_{B}\left(L\right):\,L\subseteq\mathbb{N}_{b}$,
which indicates the probability that $\left|L\right|$ targets are
born from the densities indicated by $L$. The resulting multitarget
density for new born targets is 
\begin{align}
\beta_{\tau}\left(\left\{ x_{1},...,x_{n}\right\} \right) & =\sum_{l_{1:n}}^{\neq}w_{B}\left(\left\{ l_{1},...,l_{n}\right\} \right)\prod_{j=1}^{n}\beta_{l_{j}}\left(x_{j}\right),\label{eq:birth_model_radar}
\end{align}
where $n\leq b$ and the sum is performed over distinct elements:
\begin{align*}
\sum_{l_{1:n}}^{\neq} & =\sum_{l_{1:n}:l_{1}\neq...\neq l_{n}}.
\end{align*}
The birth model in (\ref{eq:birth_model_radar}) corresponds to an
$\mathrm{MBM}_{01}$, see Appendix \ref{sec:Appendix_conjugate_prior}.
Note that we can draw samples from (\ref{eq:birth_model_radar}) by
first generating the auxiliary set $L$, whose cardinality is the
number of new born targets, from $w_{B}\left(\cdot\right)$ and then
drawing the target states from the corresponding densities independently. 
\begin{example}
Let us consider one-dimensional targets that are born according to
the model (\ref{eq:birth_model_radar}) with $b=2$,
\begin{align*}
\beta_{i}\left(x\right) & =\mathcal{N}\left(x;\mu_{i},\sigma_{i}^{2}\right)
\end{align*}
where $\mu_{i}$ and $\sigma_{i}^{2}$ represent the mean and variance
of the $i$th birth component, $w_{B}\left(\emptyset\right)=0.8$,
$w_{B}\left(\left\{ 1\right\} \right)=0.1$, $w_{B}\left(\left\{ 2\right\} \right)=0.05$
and $w_{B}\left(\left\{ 1,2\right\} \right)=0.05$. This means that
no target is born with probability 0.8, one target is born from density
$\beta_{1}\left(\cdot\right)$ with probability 0.1 and from $\beta_{2}\left(\cdot\right)$
with probability 0.05. Finally, two targets are born with probability
0.05, one from density $\beta_{1}\left(\cdot\right)$ and another
from $\beta_{2}\left(\cdot\right)$. $\oblong$ 
\end{example}
If the likelihood is (\ref{eq:likelihood_radar}) and the multitarget
density of new born targets is (\ref{eq:birth_model_radar}), we show
in this section that the multitrajectory filtering density can be
written as
\begin{align}
\pi^{k}\left(\left\{ X_{1},...,X_{n}\right\} \right) & =\sum_{h_{1:n}^{k|k}}^{\neq}w^{k|k}\left(\left\{ h_{1}^{k|k},...,h_{n}^{k|k}\right\} \right)\nonumber \\
 & \quad\times\prod_{j=1}^{n}p^{k|k}\left(X_{j}|h_{j}^{k|k}\right)\label{eq:posterior_radar}
\end{align}
where $h_{j}^{k|k}=\left(l_{j},t_{j},i_{j},\xi_{j}\right)$ is a single
trajectory hypothesis which implies that the density $p^{k|k}\left(\cdot|h_{j}^{k|k}\right)$
(on $T_{\left(k'\right)}$) has been obtained by propagating birth
component $\beta_{l_{j}}\left(\cdot\right)$ with starting time $t_{j}$,
duration $i_{j}$ and data associations $\xi_{j}$. Here, $\xi_{j}$
is a vector of length $i_{j}$ that takes values 0 if the density
$p^{k|k}\left(\cdot|h_{j}^{k|k}\right)$ is associated with clutter
or $i$ if it is associated with the $i$th measurement at the corresponding
time step. The sum in (\ref{eq:posterior_radar}) goes over over all
trajectory hypotheses that may occur jointly up to the current time.
As the birth model (\ref{eq:birth_model_radar}), this multitrajectory
density is also an $\mathrm{MBM}_{01}$. We can see that hypothesis
$h_{j}^{k|k}$ includes the pair $\left(l_{j},t_{j}\right)$, which
corresponds to the label  in \cite{Vo13}, but the label is not included
in the trajectory state. More details about the labelled approach
are given in Section \ref{subsec:Labelled-set-of-trajectories}. 

We also show that the multitrajectory predicted density at time $k$
has the same form as (\ref{eq:posterior_radar}) and can be written
as
\begin{align}
\pi^{k|k-1}\left(\left\{ X_{1},...,X_{n}\right\} \right) & =\sum_{h_{1:n}^{k|k-1}}^{\neq}w^{k|k-1}\left(\left\{ h_{1}^{k|k-1},...,h_{n}^{k|k-1}\right\} \right)\nonumber \\
 & \quad\times\prod_{j=1}^{n}p^{k|k-1}\left(X_{j}|h_{j}^{k|k-1}\right)\label{eq:prior_radar}
\end{align}
where $h_{j}^{k|k-1}$ is the same as $h_{j}^{k|k}$ but the data
association vector has $i_{j}-1$ components as the data association
at time $k$ has not been done yet. The resulting steps of the trajectory
$\mathrm{MBM}_{01}$ filtering recursion, which are explained in the
rest of the section, are shown in Procedure \ref{alg:Recursion-radar_measurement}.
As we consider that trajectories cannot be born before time step 1,
see Section \ref{subsec:State-variables}, we can set $w^{0|0}\left(\emptyset\right)=1$
such that trajectories at time 1 are born according to the birth model.
In addition, $w^{0|0}\left(\emptyset\right)=1$ is a particular case
of (\ref{eq:posterior_radar}) so the multitrajectory $\mathrm{MBM}_{01}$
density is conjugate for the standard model. 

\begin{algorithm}
\caption{\label{alg:Recursion-radar_measurement}Steps of the trajectory $\mathrm{MBM}_{01}$
filter}

{\fontsize{9}{9}\selectfont

\begin{algorithmic}     

\State - Initialisation: $w^{0|0}\left(\emptyset\right)=1$. 

\For{ $k=1$ to \emph{final time step} } 

\State - Prediction: Generate the new hypothesis sets and calculate/approximate
their weights $w^{k|k-1}\left(\cdot\right)$ and trajectory densities
$p^{k|k-1}\left(\cdot|\cdot\right)$ using Lemma \ref{lem:Prediction_radar}.

\State - Update: Generate the new hypothesis sets and calculate/approximate
their weights $w^{k|k}\left(\cdot\right)$ and trajectory densities
$p^{k|k}\left(\cdot|\cdot\right)$ using Lemma \ref{lem:Update_radar}.

\EndFor

\end{algorithmic}

}
\end{algorithm}

Before providing the recursive formulas for computing (\ref{eq:posterior_radar})
and (\ref{eq:prior_radar}), we introduce the following sets of single
trajectory hypotheses $\left(l,t,i,\xi\right)$: $\mathbb{U}^{k}$
contains the hypotheses of a present trajectory at time $k$ that
has a data association hypothesis at time $k$; $\mathbb{S}^{k}$
contains the hypotheses of a surviving trajectory at time $k$ that
does not yet contain a data association hypothesis at time $k$, $\mathbb{D}^{k}$
contains the hypotheses of trajectories present at time $k-1$ but
not present at time $k$, $\mathbb{N}^{k}$ contains the hypotheses
of new born trajectories at time $k$ and $\mathbb{D}^{1:k}=\uplus_{j=1}^{k}\mathbb{D}^{j}$,
which considers trajectories that ended at time $k-1$ or earlier.
After the $k$th update step, a single trajectory hypothesis is contained
in $\mathbb{U}^{k}\uplus\mathbb{D}^{1:k}$. Before the $k$th update
step, a single trajectory hypothesis is contained in $\mathbb{S}^{k}\uplus\mathbb{N}^{k}\uplus\mathbb{D}^{1:k}$.
Mathematically, these sets are given by 
\begin{align*}
\mathbb{U}^{k} & =\left\{ \left(l,t,i,\xi\right):l\in\mathbb{N}_{b},t\leq k,t+i-1=k,\mathrm{d}\left(\xi\right)=i\right\} \\
\mathbb{S}^{k} & =\left\{ \left(l,t,i,\xi\right):l\in\mathbb{N}_{b},t<k,t+i-1=k,\mathrm{d}\left(\xi\right)=i-1\right\} \\
\mathbb{D}^{k} & =\left\{ \left(l,t,i,\xi\right):l\in\mathbb{N}_{b},t<k,t+i=k,\mathrm{d}\left(\xi\right)=i\right\} \\
\mathbb{N}^{k} & =\left\{ \left(l,k,1\right):l\in\mathbb{N}_{b}\right\} 
\end{align*}
where $\mathrm{d}\left(\xi\right)$ denotes the dimension of vector
$\xi$.
\begin{lem}[Prediction]
\label{lem:Prediction_radar}Given $\pi^{k-1}\left(\cdot\right)$
of the form (\ref{eq:posterior_radar}) and hypothesis sets $\mathcal{A}\subset\mathbb{S}^{k}$,
$\mathcal{B}\subset\mathbb{D}^{k}$, $\mathcal{C}\subset\mathbb{N}^{k}$
and $\mathcal{D}\subset\mathbb{D}^{1:k-1}$, the predicted weight
in (\ref{eq:prior_radar}) for the hypothesis set $\mathcal{A}\uplus\mathcal{B}\uplus\mathcal{C}\uplus\mathcal{D}$
is
\begin{align}
 & w^{k|k-1}\left(\mathcal{A}\uplus\mathcal{B}\uplus\mathcal{C}\uplus\mathcal{D}\right)\nonumber \\
 & \quad=\left[\gamma_{S}\right]^{\mathcal{A}}\left[\gamma_{D}\right]^{\mathcal{B}}w_{B}^{k}\left(\mathcal{C}\right)w^{k-1|k-1}\left(\mathcal{A}^{-}\uplus\mathcal{B}\uplus\mathcal{D}\right)\label{eq:final_weight_prior}
\end{align}
\begin{align}
\gamma_{S}\left(h\right) & =\int p_{S}\left(X^{|}\right)p^{k-1|k-1}\left(X|h^{-}\right)dX\label{eq:survival_weight}\\
\gamma_{D}\left(h\right) & =1-\gamma_{S}\left(h\right)\label{eq:dead_weight}
\end{align}
\begin{align}
w_{B}^{k}\left(\left\{ \left(l_{1},k,1\right),...,\left(l_{n},k,1\right)\right\} \right) & =w_{B}\left(\left\{ l_{1},...,l_{n}\right\} \right)\label{eq:new_born_weight}
\end{align}
where for $h=\left(l,t,i,\xi\right)$, we have $h^{-}=\left(l,t,i-1,\xi\right)$,
$\mathcal{A}^{-}=\left\{ h^{-}:h\in\mathcal{A}\right\} $ and $X^{|}$
is the last target state of $X$. Also, $w_{B}^{k}\left(\cdot\right)$
is zero if evaluated at global hypotheses different from (\ref{eq:new_born_weight}).

The single trajectory density for $h\in\mathbb{S}^{k}\uplus\mathbb{D}^{k}\uplus\mathbb{N}^{k}\uplus\mathbb{D}^{1:k-1}$
is
\begin{align}
 & p^{k|k-1}\left(X|h\right)\nonumber \\
 & =p_{S}^{k|k-1}\left(X|h\right)1_{\mathbb{S}^{k}}\left(h\right)+p_{D}^{k|k-1}\left(X|h\right)1_{\mathbb{D}^{k}}\left(h\right)\nonumber \\
 & \quad+p_{N}^{k|k-1}\left(X|h\right)1_{\mathbb{N}^{k}}\left(h\right)+p^{k-1|k-1}\left(X|h\right)1_{\mathbb{D}^{1:k-1}}\left(h\right)\label{eq:PDF_trajectory_given_hypothesis_prior}
\end{align}
where
\begin{align*}
p_{S}^{k|k-1}\left(t,x^{1:i}|h\right) & =g\left(x^{i}|x^{i-1}\right)p_{S}\left(x^{i-1}\right)\\
 & \quad\times p^{k-1|k-1}\left(t,x^{1:i-1}|h^{-}\right)/\gamma_{S}\left(h\right)\\
p_{D}^{k|k-1}\left(t,x^{1:i}|h\right) & =\frac{p^{k-1|k-1}\left(t,x^{1:i}|h\right)\left(1-p_{S}\left(x^{i}\right)\right)}{\gamma_{D}\left(h\right)}\\
p_{N}^{k|k-1}\left(t,x^{1}|l,k,1\right) & =\beta_{l}\left(x^{1}\right)\delta_{k}\left[t\right].
\end{align*}
\end{lem}
This lemma is proved in Appendix \ref{sec:Appendix_conjugate_prior}.
Each hypothesis set in (\ref{eq:final_weight_prior}) can be decomposed
into disjoint hypothesis sets $\mathcal{A}$, $\mathcal{B}$, $\mathcal{C}$
and $\mathcal{D}$ that describe surviving trajectories, present trajectories
at time $k-1$ but not at time $k$, new born trajectories and trajectories
that were present some time before time $k-1$ but not at time $k-1$,
respectively. The weight $w^{k|k-1}\left(\cdot\right)$ corresponds
to the weight $w^{k-1|k-1}\left(\cdot\right)$ of the parent hypothesis
set $\mathcal{A}^{-}\uplus\mathcal{B}\uplus\mathcal{D}$ multiplied
by the weight of the hypothesis set $\mathcal{C}$ of new born targets,
the probability (\ref{eq:survival_weight}) of survival for trajectories
hypothesised in $\mathcal{A}$ and the probability (\ref{eq:dead_weight})
of death for trajectories hypothesised in $\mathcal{B}$. The resulting
density of a trajectory given a hypothesis is given by (\ref{eq:PDF_trajectory_given_hypothesis_prior}).
Densities $p_{S}^{k|k-1}\left(\cdot|h\right)$, $p_{D}^{k|k-1}\left(\cdot|h\right)$
and $p_{N}^{k|k-1}\left(\cdot|h\right)$ correspond to a surviving
trajectory, a trajectory present at time $k-1$ but not at time $k$,
and a new born trajectory, respectively. If the trajectory is not
present at time $k-1$, which means that its hypothesis is contained
in $\mathbb{D}^{1:k-1}$, its density remains unaltered. 
\begin{lem}[Update]
\label{lem:Update_radar}Given $\pi^{k|k-1}\left(\cdot\right)$ of
the form (\ref{eq:prior_radar}), the measurement set $\mathbf{z}^{k}=\left\{ z_{1}^{k},...,z_{m}^{k}\right\} $
and hypothesis sets $\mathcal{D}\subset\mathbb{D}^{1:k}$ and $\mathcal{E}=\left\{ h_{1}^{k|k},...,h_{n}^{k|k}\right\} \subset\mathbb{U}^{k}$
such that $h_{j}^{k|k}=\left(h_{j}^{k|k-1},\theta_{j}\right)$, $\left(\theta_{1},...,\theta_{n}\right)\in\Theta_{n,m}$,
the filtering weight in (\ref{eq:posterior_radar}) for hypothesis
set $\mathcal{E}\uplus\mathcal{D}$ is
\begin{align}
w^{k|k}\left(\mathcal{E}\uplus\mathcal{D}\right) & \propto w^{k|k-1}\left(\mathcal{E}^{\circ}\uplus\mathcal{D}\right)\left[\eta_{\mathbf{z}^{k}}\right]^{\mathcal{E}}\label{eq:final_weight_update}\\
\eta_{\mathbf{z}^{k}}\left(h_{j}^{k|k-1},\theta_{j}\right) & =\int\psi_{\mathbf{z}^{k}}\left(X|\theta_{j}\right)p^{k|k-1}\left(X|h_{j}^{k|k-1}\right)dX
\end{align}
where $\mathcal{E}^{\circ}=\left\{ h_{1}^{k|k-1},...,h_{n}^{k|k-1}\right\} $.

The single trajectory density for $h_{j}^{k|k}\in\mathbb{U}^{k}\uplus\mathbb{D}^{1:k}$
is
\begin{align}
p^{k|k}\left(X|h_{j}^{k|k}\right) & =p_{U}^{k|k}\left(X|h_{j}^{k|k}\right)1_{\mathbb{U}^{k}}\left(h_{j}^{k|k}\right)\nonumber \\
 & \quad+p^{k|k-1}\left(X|h_{j}^{k|k}\right)1_{\mathbb{D}^{1:k}}\left(h_{j}^{k|k}\right)\label{eq:PDF_trajectory_given_hypothesis_posterior}\\
p_{U}^{k|k}\left(X|h_{j}^{k|k-1},\theta_{j}\right) & =\frac{\psi_{\mathbf{z}^{k}}\left(X|\theta_{j}\right)p^{k|k-1}\left(X|h_{j}^{k|k-1}\right)}{\eta_{\mathbf{z}^{k}}\left(h_{j}^{k|k-1},\theta_{j}\right)}.\nonumber 
\end{align}
\end{lem}
This lemma is proved in Appendix \ref{sec:Appendix_conjugate_prior}.
A hypothesis set in (\ref{eq:final_weight_update}) can be decomposed
into disjoint hypothesis sets $\mathcal{E}\uplus\mathcal{D}$ that
describe present trajectories at time $k$ (with a data association
hypothesis at time $k$) and trajectories that were present before
time $k$ but not at time $k$, respectively. The weight $w^{k|k}\left(\cdot\right)$
corresponds to the weight $w^{k|k-1}\left(\cdot\right)$ of the parent
hypothesis $\mathcal{E}^{\circ}\uplus\mathcal{D}$ multiplied by the
data association probabilities $\eta_{\mathbf{z}^{k}}\left(\cdot\right)$
of the present trajectories. The resulting density of a trajectory
for a hypothesis is given by (\ref{eq:PDF_trajectory_given_hypothesis_posterior}).
Density $p_{U}^{k|k}\left(\cdot|h_{j}^{k|k}\right)$ corresponds to
a present trajectory at time $k$. If the trajectory is not present
at time $k$, which means that its hypothesis is contained in $\mathbb{D}^{1:k}$,
its density remains unaltered. 

In the trajectory $\mathrm{MBM}_{01}$ filter, it is important to
highlight that the update and prediction of the weights only depend
on the densities of the target states at the current time. In other
words, $\gamma_{S}\left(\cdot\right)$, $\gamma_{D}\left(\cdot\right)$
and $\eta_{\mathbf{z}^{k}}\left(\cdot\right)$ are calculated/approximated
using a single target integral w.r.t. the density of the target $\tau^{k}\left(X\right)$
for the corresponding hypothesis. For linear/Gaussian dynamic and
measurement models, with constant $p_{D}\left(\cdot\right)$ and $p_{S}\left(\cdot\right)$,
and Gaussian $\beta_{i}\left(\cdot\right)$, Lemmas \ref{lem:Prediction_radar}
and \ref{lem:Update_radar} can be implemented in closed-form, subject
to the practical considerations of managing an ever-increasing number
of hypotheses and trajectories, which are discussed in Section \ref{subsec:Practical-considerations}.
The resulting formulas can be found in Appendix \ref{sec:Appendix_Gaussian_implementation}
and are illustrated via simulations in Section \ref{sec:Illustrative-example}.
If the system is nonlinear/non-Gaussian, we need to perform single
trajectory density approximations to calculate (\ref{eq:PDF_trajectory_given_hypothesis_prior})
and (\ref{eq:PDF_trajectory_given_hypothesis_posterior}). 

We would also like to point out that there are three types of conjugate
priors for unlabelled multi-target filtering in the literature, which
depend on the birth model and the structure of the hypotheses \cite[Sec. IV]{Angel18_b}:
$\mathrm{MBM}_{01}$, multi-Bernoulli mixture (MBM), and Poisson multi-Bernoulli
mixture (PMBM). This paper has introduced the $\mathrm{MBM}_{01}$
conjugate prior for sets of trajectories. Conjugacy for sets of trajectories
also holds for PMBMs \cite{Granstrom18}, and for MBMs, which are
a particular case of PMBMs \cite[Sec. III.E]{Angel18_b}. One of the
advantages of the PMBM conjugate prior is that information on non-detected
trajectories is represented efficiently with the Poisson component.

\subsection{Practical considerations\label{subsec:Practical-considerations}}

We want to recall that the main purpose of this paper is to establish
the foundations to perform MTT using sets of trajectories, not the
development of efficient, practical algorithms. Nevertheless, in this
section, we discuss some practical considerations for algorithms based
on set of trajectories. 

It should be noted that we cannot run the trajectory $\mathrm{MBM}_{01}$
filtering recursion in Procedure \ref{alg:Recursion-radar_measurement}
for a long time without approximations due to a linearly increasing
state dimension and a super-exponentially increasing number of hypotheses.
The problem of managing an ever increasing number of hypotheses can
be addressed by using gating or pruning using Murty's algorithm \cite{Murty68}
in Lemma \ref{lem:Update_radar}, as in MHT and $\delta$-GLMB filter.
Another possibility, widely used in MHT, is to prune hypotheses considering
the data association over multiple scans jointly \cite{Blackman04}.
We would like to clarify that pruning consists of approximating some
of the weights in the $\mathrm{MBM}_{01}$ that represents the filtering
(multitrajectory) density (\ref{eq:posterior_radar})  as zero, followed
by a normalisation of the resulting weights. Therefore, pruning does
not affect the symmetry of the distribution. 

A practical approach to deal with densities of states with increasing
dimensionality is explained in \cite{Angel18_c} for the trajectory
PHD filter. The main idea is to only update the single trajectory
densities over a time window that contains the last few time steps
as, in practice, measurements at the current time step only have a
significant impact on the trajectory state for recent time steps.
Another option is to just consider the distribution of the sets of
trajectories over a sliding time window. For example, we can process
each single trajectory density as in the accumulated state densities
in \cite{Koch11}, or we can just consider the distribution of the
set of trajectories over the last two time steps as done in \cite[Sec. IV.B]{Angel14}
to estimate trajectories sequentially for fixed and known number of
targets. Using these practical considerations, we present an online
algorithm for MTT using sets of trajectories in Section \ref{sec:Illustrative-example}.

Finally, we would like to comment on trajectory estimation, though
it is not the main topic of this paper. With the multitrajectory density,
we can estimate the best possible trajectories up to the current time
step, following a certain criterion. For example, we can choose the
global hypothesis with highest weight and report the corresponding
posterior mean of the trajectories, see Section \ref{sec:Illustrative-example},
or we can, in principle, use an estimate that minimises the posterior
expected loss according to a metric for sets of trajectories. Another
approach to estimation is to estimate trajectories sequentially. That
is, we append new target estimates at the current time step to estimated
trajectories at the previous time step. It should be noted that sequential
track builders do not use all available information to estimate the
trajectories up to the current time and the resulting estimates do
not necessarily represent reasonable trajectories. For example, two
estimated labelled target states at two consecutive time steps based
on the global hypotheses with highest weight do not necessarily represent
good trajectories, as the underlying data association hypothesis in
each global hypothesis can be significantly different. Nevertheless,
the choice between sequential track estimators and non-sequential
track estimators can be part of the problem formulation, and both
can be tackled with sets of trajectories or sequences of labelled
sets.

\section{Relation with other MTT models\label{sec:Relations-with-other}}

In this section we relate the proposed filtering based on set of trajectories
with other labelled and unlabelled RFS models and classical MHT. 

\subsection{Relation with sets of labelled trajectories\label{subsec:Labelled-set-of-trajectories}}

As explained above, MTT of targets without identification can be performed
without labels. Nevertheless, in this section, we analyse the labelling
of the trajectories to establish a link with the labelled approach,
which was discussed in Section \ref{subsec:Motivation-for-sets}.
In \cite{Vo13}, a target label is the pair $\left(t,l\right)$, where
$t$ is the birth time and $l$ is the component of the birth model
(\ref{eq:birth_model_radar}) from which the target is generated.
A labelled target state corresponds to adding this (unique) label
to its state. 

While the literature has focused on sets of labelled targets and sequences
of sets of labelled targets, the approach can be easily extended to
sets of labelled trajectories. In this case, a labelled trajectory
state is formed as $\left(l,t,x^{1:i}\right)$ where $l$ indicates
that it was born from component $l$ of the birth model, see (\ref{eq:birth_model_radar}).

In this case, for the birth model (\ref{eq:birth_model_radar}) and
the standard measurement model, the calculation of the labelled multitrajectory
filtering density $\pi_{\ell}^{k}\left(\cdot\right)$ is analogous
to what was presented previously in (\ref{eq:posterior_radar}):
\begin{align}
 & \pi_{\ell}^{k}\left(\left\{ \left(l'_{1},X_{1}\right),...,\left(l'_{n},X_{n}\right)\right\} \right)\nonumber \\
 & \,=\sum_{h_{1:n}^{k|k}}^{\neq}w^{k|k}\left(\left\{ h_{1}^{k|k},...,h_{n}^{k|k}\right\} \right)\prod_{j=1}^{n}p_{\ell}^{k|k}\left(\left(l'_{j},X_{j}\right)\left|h_{j}^{k|k}\right.\right)\label{eq:labelled_unlabelled_relation}
\end{align}
where
\begin{align}
p_{\ell}^{k|k}\left(\left(l',X\right)|l,t,i,\xi\right) & =\delta_{l}\left[l'\right]p^{k|k}\left(X|l,t,i,\xi\right).\label{eq:labelled_no_labelled_equivalence}
\end{align}
That is, compared to (\ref{eq:posterior_radar}), the labelled multitrajectory
filtering density simply consists of adding a Kronecker delta determined
by the initial birth component $l$ in the single trajectory hypothesis,
which is formed by $\left(l,t,i,\xi\right)$. Therefore, in MTT with
the birth model (\ref{eq:birth_model_radar}) and standard measurement
model, variable $l$ does not form part of the state for sets of unlabelled
trajectories but is incorporated into the trajectory state to form
label $\left(t,l\right)$ for sets of labelled trajectories. The labelled
version of the trajectory $\mathrm{MBM}_{01}$ filter is referred
to as the labelled trajectory $\mathrm{MBM}_{01}$ filter and, as
indicated above, it does not imply changes in the recursion, only
in the trajectory state. 

We can make a direct equivalence between the birth model (\ref{eq:birth_model_radar})
and the $\delta$-GLMB birth model \cite[Eq. (26)]{Vo13}, by adding
explicit time dependence on the birth model (\ref{eq:birth_model_radar}),
which yields $w_{B}^{k}\left(\cdot\right)$ and $\beta_{l_{j}}^{k}\left(\cdot\right)$,
and labelling this multi-target density. Then, the birth multi-target
density becomes
\begin{align}
 & \beta_{\tau,l}^{k}\left(\left\{ \left(l'_{1},t_{1},x_{1}\right),...,\left(l'_{n},t_{n},x_{n}\right)\right\} \right)\nonumber \\
 & =\left[\prod_{j=1}^{n}\delta_{k}\left[t_{j}\right]\right]\sum_{l_{1:n}}^{\neq}w_{B}^{k}\left(\left\{ l_{1},...,l_{n}\right\} \right)\prod_{j=1}^{n}\left(\delta_{l_{j}}\left[l_{j}'\right]\beta_{l_{j}}^{k}\left(x_{j}\right)\right)\nonumber \\
 & =\left[\prod_{j=1}^{n}\delta_{k}\left[t_{j}\right]\right]w_{B}^{k}\left(\left\{ l'_{1},...,l'_{n}\right\} \right)\prod_{j=1}^{n}\beta_{l'_{j}}^{k}\left(x_{j}\right)\label{eq:labelled_birth_model}
\end{align}
if $l'_{1}\neq...\neq l'_{n}$ and zero otherwise. This birth model
is the same as the $\delta$-GLMB birth model \cite[Eq. (26)]{Vo13},
which is an $\mathrm{MBM}_{01}$ (see Section \ref{subsec:Conjugate-form})
with uniquely labelled targets. 

Note that the set of labelled targets at time $k$ can be simply obtained
by obtaining the corresponding target state at time $k$ and its label
form the labelled trajectories, as was done in (\ref{eq:set_targets_formula}).
Equivalently, there is a mapping from the set of labelled trajectories
up to time step $k$ to a sequence of sets of labelled targets so
we can obtain the same information from both representations. However,
to our knowledge, a formula such as (\ref{eq:labelled_unlabelled_relation}),
which is written in terms of single trajectory densities and is closed-form
for linear/Gaussian systems, has not been proposed yet for sequence
of labelled sets.

\subsection{Relation with MHT algorithms}

Classical MHT algorithms \cite{Reid79,Mori86,Kurien_inbook90} were
developed for the standard measurement model but not for general track-before-detect
models \cite{Angel16}. They rely on enumerating multiple target/measurement
association hypotheses, calculating their probabilities and the density
of the current target state given a hypothesis. For the standard measurement
model, there are also algorithms based on RFS of targets that have
this MHT-type structure \cite{Williams15b,Vo13,Mahler_book14}. Moreover,
for this measurement model, the multitrajectory filtering density,
which is given by (\ref{eq:posterior_radar}), is actually also computed
by considering multiple hypotheses but with densities over trajectories
instead of targets. This is analogous to the algorithms in \cite{Koch11},
which assume a single, always existing trajectory. In this sense,
the trajectory $\mathrm{MBM}_{01}$ filter can be considered a form
of MHT algorithm, derived using sets of trajectories and FISST. 

Compared to classical MHT and MHT-type algorithms for RFS of targets,
we introduce the set of trajectories as a random variable that represents
all quantities of interest. This enables us to compute the posterior
distribution of the set of trajectories in a direct Bayesian manner
and answer all types of questions regarding the history, as exemplified
in Section \ref{subsec:Motivation-for-sets}. Another advantage of
introducing sets of trajectories is that we can use definitions of
estimation error based on metrics \cite{Rahmathullah16_prov2,Bento_draft16}.
In addition, compared to classical MHT algorithms, sets of trajectories
can also be used with other measurement models, such as track-before-detect,
or to develop new algorithms, such as the trajectory PHD filter \cite{Angel18_c},
which do not give rise to an MHT-type structure. Nevertheless, the
development of efficient MHT-type algorithms based on sets of trajectories
should be inspired by the vast literature on classical MHT and MHT-type
algorithms with RFS of targets. 

\subsection{Relation with sets of targets}

\begin{figure}
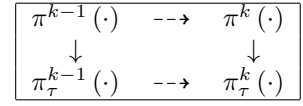

\begin{centering}
\begin{tabular}{|ccc|}
\hline 
$\pi^{k-1}\left(\cdot\right)$ &
$\dashrightarrow$ &
$\pi^{k}\left(\cdot\right)$\tabularnewline
$\downarrow$ &
 &
$\downarrow$\tabularnewline
$\pi_{\tau}^{k-1}\left(\cdot\right)$ &
$\dashrightarrow$ &
$\pi_{\tau}^{k}\left(\cdot\right)$\tabularnewline
\hline 
\end{tabular}
\par\end{centering}
\caption{\label{fig:Relation-marginalisation-prediction-update}Relation between
the multitarget filtering densities $\pi_{\tau}^{k-1}\left(\cdot\right)$
and $\pi_{\tau}^{k}\left(\cdot\right)$ and the multitrajectory filtering
densities $\pi^{k-1}\left(\cdot\right)$ and $\pi^{k}\left(\cdot\right)$.
Dashed arrows stand for prediction and update and the solid arrows
for marginalisation.}
\end{figure}

This section relates the proposed approach to the multitarget filtering
approach based on sets of targets. We first provide a theorem for
obtaining the multitarget density of the set of targets at a particular
time given a multitrajectory density. This process resembles marginalisation
of densities in vector spaces \cite{Sarkka_book13}. 
\begin{thm}
\label{thm:Marginalisation}Given the multitrajectory density $\pi\left(\cdot\right)$
of $\mathbf{X}$, the multitarget filtering density $\pi_{\tau}^{k}\left(\cdot\right)$
of $\tau^{k}\left(\mathbf{X}\right)$ is given by
\begin{align*}
\pi_{\tau}^{k}\left(\mathbf{y}\right) & =\int\delta_{\tau^{k}\left(\mathbf{X}\right)}\left(\mathbf{y}\right)\pi\left(\mathbf{X}\right)\delta\mathbf{X}
\end{align*}
where
\begin{align*}
\delta_{\mathbf{z}}\left(\mathbf{y}\right) & =\begin{cases}
0 & \mathrm{if}\,\left|\mathbf{z}\right|\neq\left|\mathbf{y}\right|\\
1 & \mathrm{if}\,\left|\mathbf{z}\right|=\left|\mathbf{y}\right|=0\\
\sum_{\sigma\in\Gamma_{n}}\prod_{j=1}^{n}\delta_{z_{\sigma_{j}}}\left(y_{j}\right) & \mathrm{if}\,\begin{cases}
\mathbf{y} & =\left\{ y_{1},...,y_{n}\right\} \\
\mathbf{z} & =\left\{ z_{1},...,z_{n}\right\} 
\end{cases}
\end{cases}
\end{align*}
is a multitarget Dirac delta centered at $\mathbf{z}$ \cite[Eq. (11.124)]{Mahler_book07}
and $\Gamma_{n}$ is the set of all the permutations of $\left(1,...,n\right)$.
\end{thm}
Theorem \ref{thm:Marginalisation} is proved in Appendix \ref{sec:Appendix_marginalisatioin}.
In the usual RFS filtering framework with sets of targets, we obtain
the multitarget filtering density $\pi_{\tau}^{k}\left(\cdot\right)$
using the filtering density $\pi_{\tau}^{k-1}\left(\cdot\right)$
and the corresponding prediction and update equations \cite{Mahler_book07}.
In Appendix \ref{sec:Appendix_relation_usual_RFS}, we prove that
given the multitrajectory density $\pi^{k-1}\left(\cdot\right)$ and
its corresponding multitarget density $\pi_{\tau}^{k-1}\left(\cdot\right)$
at time $k-1$, if we apply Theorem \ref{thm:Marginalisation} to
the output of the prediction and update steps, which are given by
(\ref{eq:prediction_trajectories}) and (\ref{eq:update_trajectories}),
as expected, we obtain the same multitarget filtering density for
the set of targets at time $k$ as in the usual RFS filtering framework.
This is illustrated in Figure \ref{fig:Relation-marginalisation-prediction-update}.

This marginalisation result applies to both labelled and unlabelled
sets, which implies an important relation. The $\delta$-GLMB filter
recursively computes the multi-target posterior $\pi_{\tau}^{k}\left(\cdot\right)$
for labelled sets, standard dynamic and measurement models, and birth
model (\ref{eq:labelled_birth_model}). The $\mathrm{MBM}_{01}$ recursion
for sets of labelled trajectories, which is analogous to the unlabelled
recursion and yields the posterior multi-trajectory density (\ref{eq:labelled_unlabelled_relation}),
is also obtained with the same models. Therefore, the labelled multitarget
filtering density obtained by marginalisation of (\ref{eq:labelled_unlabelled_relation})
at the current time step is equivalent to the multi-target filtering
density obtained by the $\delta$-GLMB filter. We can therefore consider
the labelled trajectory $\mathrm{MBM}_{01}$ filter in Section \ref{subsec:Labelled-set-of-trajectories}
as the extension of the $\delta$-GLMB filter to labelled trajectories. 

\section{Illustrative examples\label{sec:Illustrative-example}}

This section illustrates how to perform multiple target tracking using
sets of trajectories via simulations. We first illustrate the form
of the multitrajectory density in Section \ref{subsec:Illustration-of-the-multitrajectory}.
In Section \ref{subsec:Implementation-using-Murty's}, we show how
the trajectory $\mathrm{MBM}_{01}$ filter can be implemented using
Murty's algorithm and a sliding time window. All units of this section
are in the international system.

\subsection{One dimensional scenario\label{subsec:Illustration-of-the-multitrajectory}}

We assume a target state $x\in\mathbb{R}^{2}$ that consists of position
and velocity. We only consider position in a one-dimensional space
so that we can visualise the results easily. The birth process has
parameters: $b=2$, $\beta_{1}\left(x\right)=\beta_{2}\left(x\right)=\mathcal{N}\left(x;\left[0,0\right]^{T},\mathrm{diag}\left(25,1\right)\right)$,
$w_{B}\left(\oslash\right)=0.85$, $w_{B}\left(\left\{ 1\right\} \right)=w_{B}\left(\left\{ 2\right\} \right)=0.05$,
$w_{B}\left(\left\{ 1,2\right\} \right)=0.05$. The single-target
dynamic process parameters are: $p_{S}=0.9$, $g\left(x^{i}|x^{i-1}\right)=\mathcal{N}\left(x^{i};Fx^{i-1},Q\right)$
where 
\begin{align*}
F= & \left(\begin{array}{cc}
1 & 1\\
0 & 1
\end{array}\right),\quad Q=\frac{1}{10}\left(\begin{array}{cc}
1/3 & 1/2\\
1/2 & 1
\end{array}\right).
\end{align*}
The intensity function of the Poisson clutter is $\kappa\left(z\right)=1.4\cdot\frac{1}{20}1_{\left(-10,10\right)}\left(z\right)$,
which means that clutter is uniformly distributed in $\left(-10,10\right)$
and there is an average of $1.4$ clutter measurement per scan. The
target-generated measurements have the following parameters: $p_{D}=0.95$,
$l\left(z|x\right)=\mathcal{N}\left(z;Hx,R\right)$ where $H=\left(\begin{array}{cc}
1 & 0\end{array}\right)$ and $R=10^{-4}$.  

\begin{figure}
\begin{centering}
\subfloat[]{\begin{centering}
\includegraphics[scale=0.29]{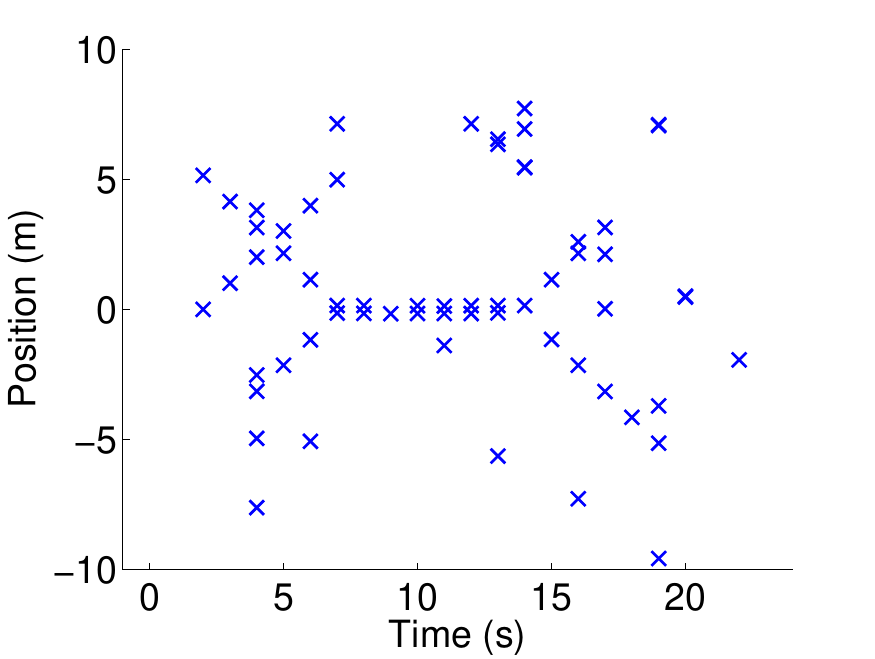}
\par\end{centering}
}\subfloat[]{\begin{centering}
\includegraphics[scale=0.29]{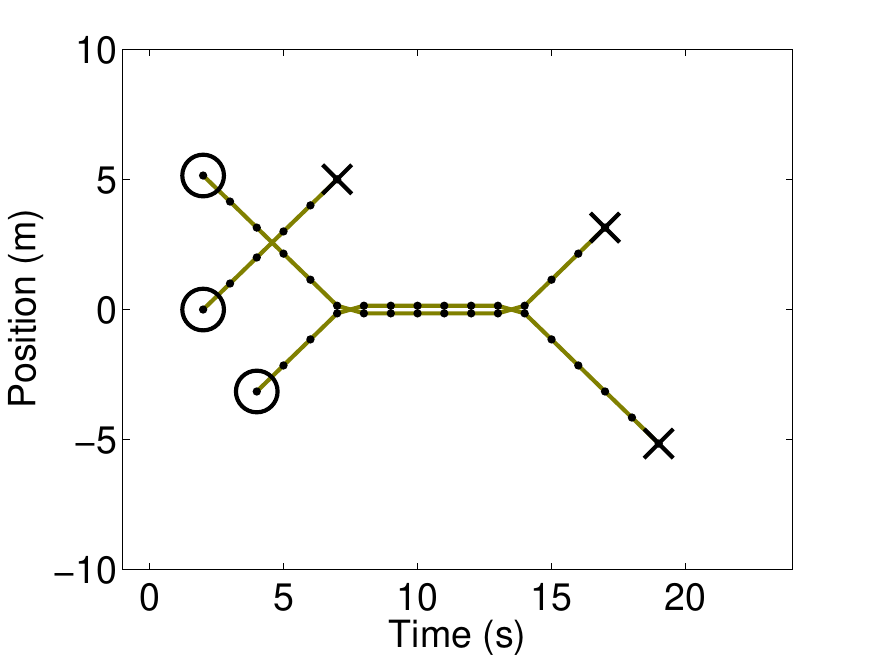}
\par\end{centering}
}
\par\end{centering}
\caption{\label{fig:Observed-measurements}Scenario of the simulation: (a)
Observed sequence of measurements. At time 14, there are two measurements
close together at positions 5.45 and 5.47. (b) Trajectories from which
the measurements have been generated. Circles and crosses indicate
the start and end of trajectories. }
\end{figure}

We consider 22 time steps and observe the measurements shown in Figure
\ref{fig:Observed-measurements}(a). These measurements have been
generated from the set of trajectories represented in Figure \ref{fig:Observed-measurements}(b).
Given these measurements and the model parameters indicated above,
we calculate the multitrajectory filtering density using Procedure
\ref{alg:Recursion-radar_measurement}. The calculations of the prediction
and update steps for the considered model are provided in Appendix
\ref{sec:Appendix_Gaussian_implementation}. In this set-up, the density
of a trajectory given a single trajectory hypothesis is Gaussian with
a certain posterior mean and covariance matrix. After each update,
we perform pruning with 100 hypotheses, which is sufficient for the
illustrating purposes of this section. 

\begin{figure}
\begin{centering}
\subfloat[0.42]{\begin{centering}
\includegraphics[scale=0.29]{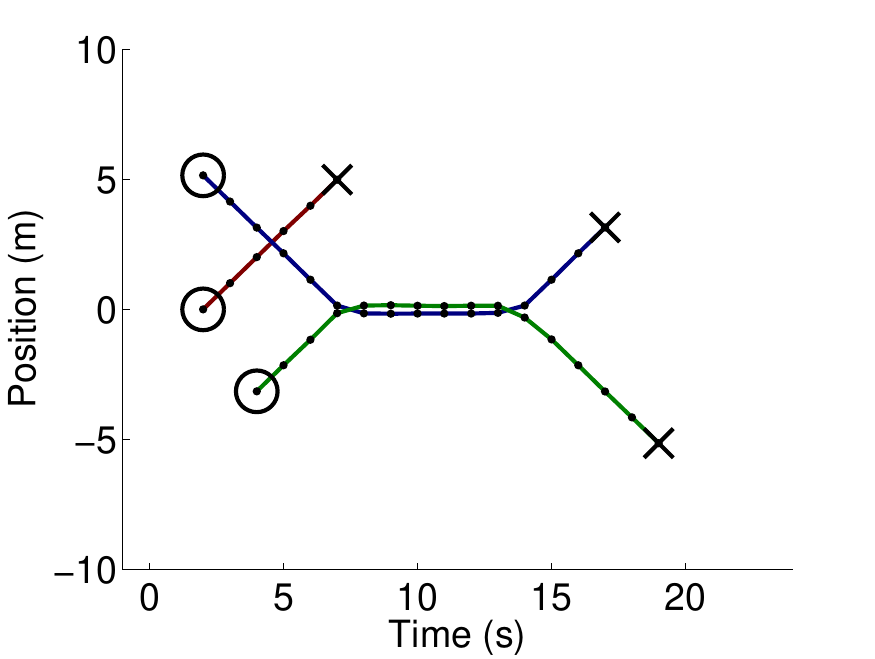}
\par\end{centering}

}\subfloat[0.12]{\begin{centering}
\includegraphics[scale=0.29]{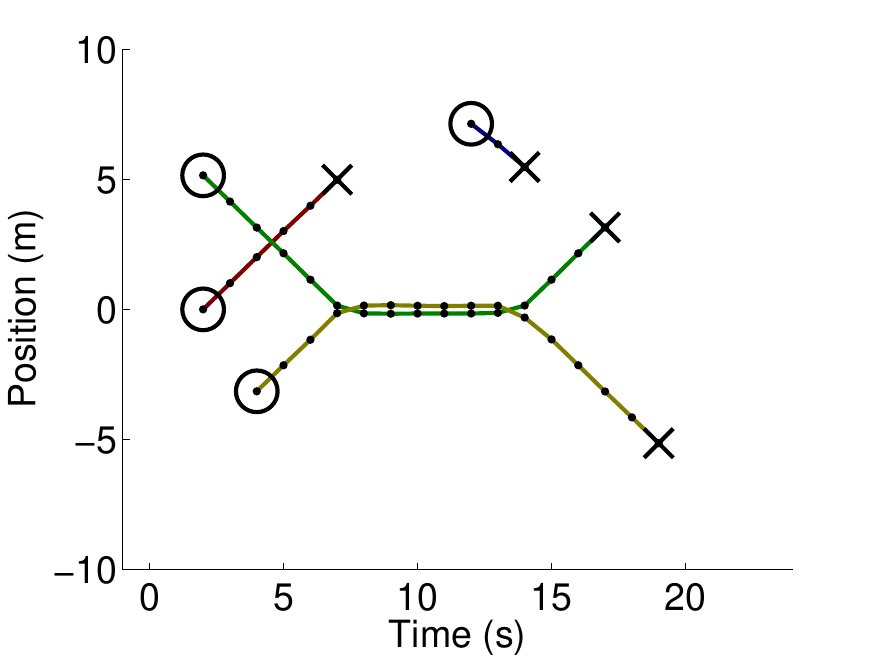}
\par\end{centering}
}
\par\end{centering}
\begin{centering}
\subfloat[0.12]{\begin{centering}
\includegraphics[scale=0.29]{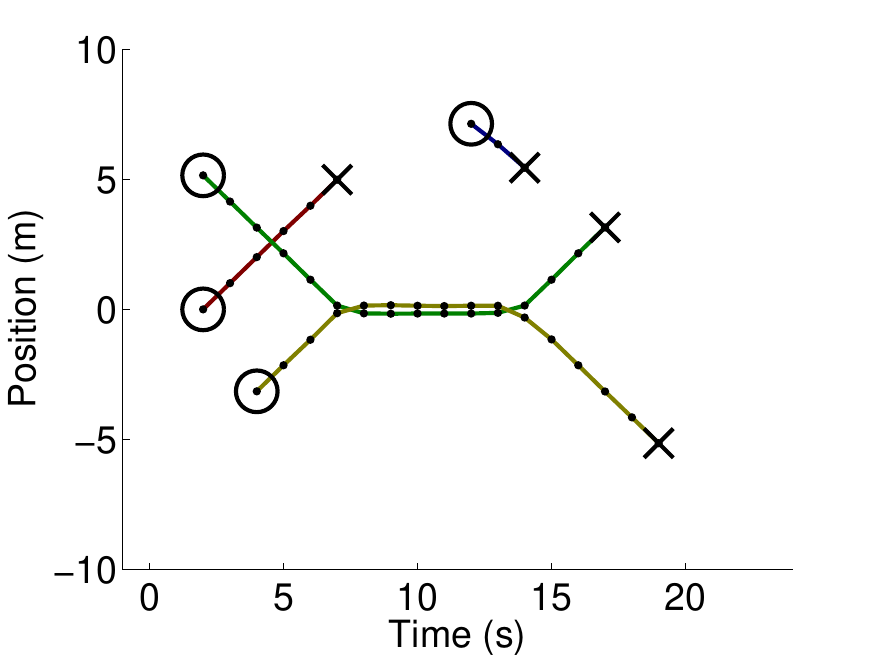}
\par\end{centering}

}\subfloat[0.04]{\begin{centering}
\includegraphics[scale=0.29]{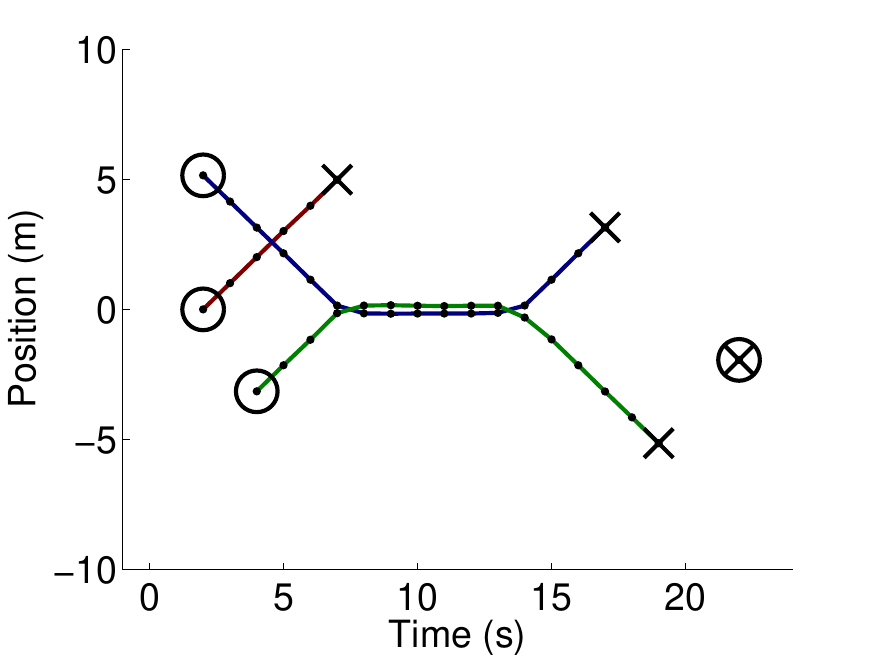}
\par\end{centering}

}
\par\end{centering}
\begin{centering}
\subfloat[0.02]{\begin{centering}
\includegraphics[scale=0.29]{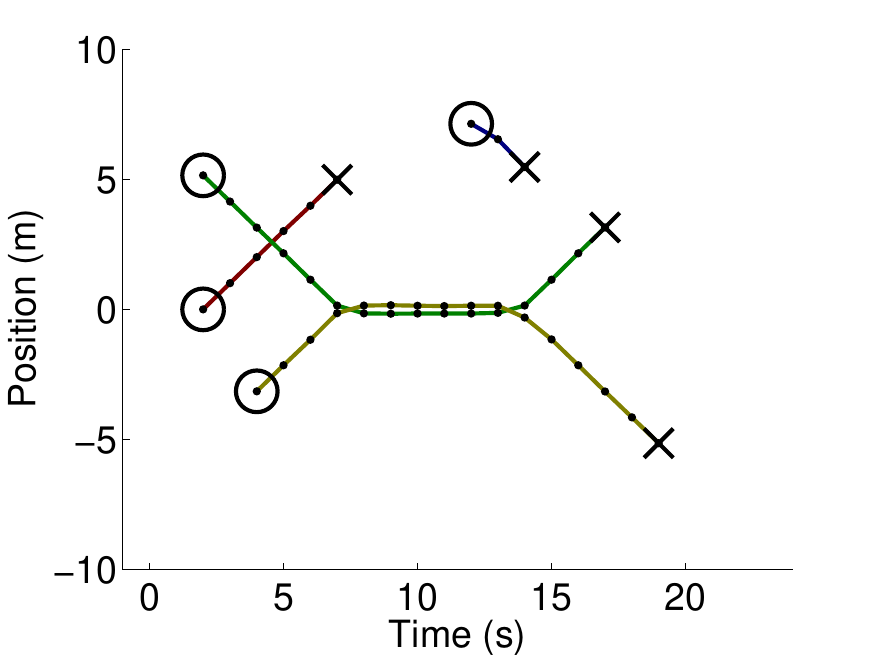}
\par\end{centering}

}\subfloat[0.02]{\begin{centering}
\includegraphics[scale=0.29]{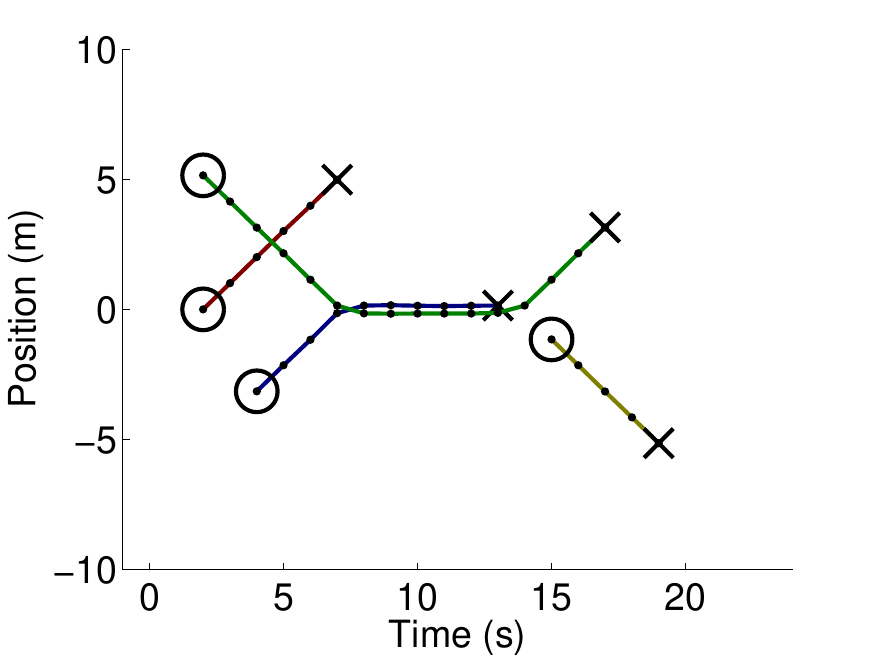}
\par\end{centering}

}
\par\end{centering}
\caption{\label{fig:Posterior-mean-trajectories}Posterior mean of the trajectories
for the six global hypotheses with largest weights, which are given
in the subfigure captions. Colors for different trajectories are only
used to help visualisation, they are not labels. }
\end{figure}

We represent the posterior mean of the trajectories for the six hypotheses
with largest weights in Figure \ref{fig:Posterior-mean-trajectories}.
The most likely situation is the one shown in Figure \ref{fig:Posterior-mean-trajectories}(a),
in which there are three trajectories. This global hypothesis accurately
represents the set of trajectories from which measurements where generated,
see Figure \ref{fig:Observed-measurements}(b). We can also see that,
according to this global hypothesis, there are two targets born at
time 2 and their trajectories are fully distinguishable: the target
at around position 0 at time 2 follows a straight line to be at around
position 5 at time step 7 while the target at around position 5 at
time 2 follows the trajectory indicated in the figure and disappears
at time step 18. In the second most likely situation, there is an
extra trajectory that appears at time 12. The third hypothesis, is
a variant of the previous with a different measurement association
at time 14 for that trajectory. Note that there are two very close
measurements at time 14 in the region where that target lies. The
fourth hypothesis includes a new born target at time step 22. The
fifth one considers a different data association at time 13 for the
trajectory that appears at time 12. The sixth hypothesis breaks one
of the trajectories and considers two trajectories instead of one.

\begin{figure}
\begin{centering}
\subfloat[]{\begin{centering}
\includegraphics[scale=0.29]{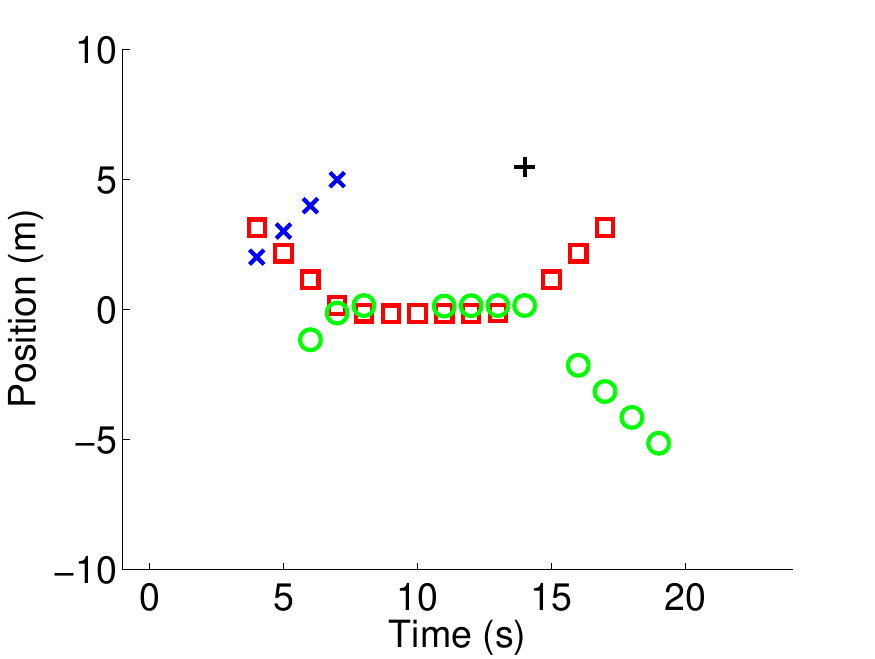}
\par\end{centering}

}\subfloat[]{\begin{centering}
\includegraphics[scale=0.29]{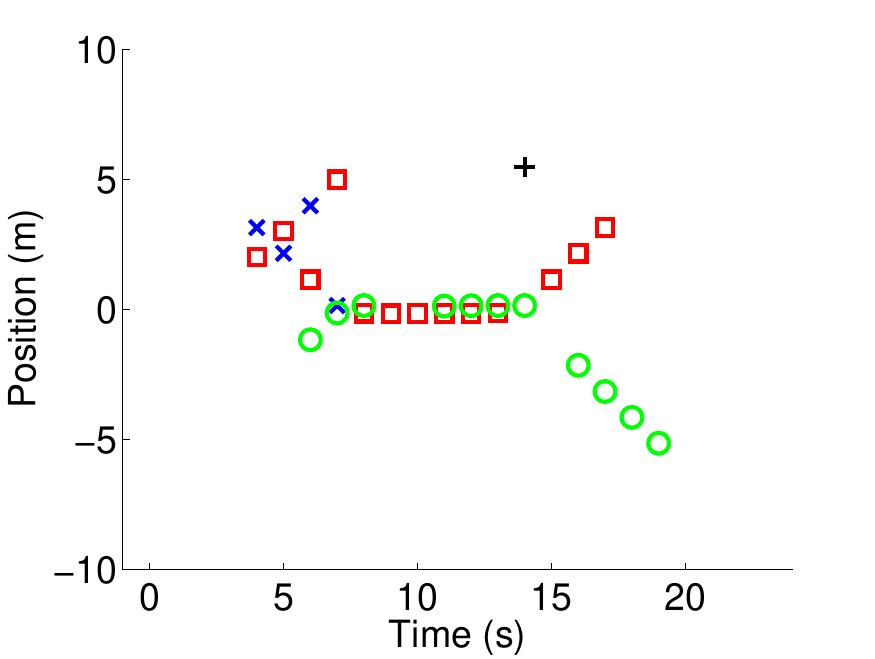}
\par\end{centering}

}
\par\end{centering}
\caption{\label{fig:Estimates_simulation_deltaGLMB}Two equally likely estimates
provided by the $\delta$-GLMB filter. Different markers represent
different labels. Track formation using multitarget filtering densities
does not work well due to track switching that can happen for targets
born at the same time. }
\end{figure}

In this set-up, it seems convenient to estimate the set of trajectories
as the posterior mean of the hypothesis with highest weight, which
is represented by Figure \ref{fig:Posterior-mean-trajectories}(a)
and does not have track switching. We also want to illustrate the
estimated trajectories using filtering multitarget densities to illustrate
the problem of track switching by implementing the $\delta$-GLMB
filter \cite{Vo13}. We would like to remark that the $\delta$-GLMB
filter only computes the multitarget density at the current time so
it is considerably lighter than the direct implementation of the conjugate
prior for sets of trajectories in Procedure \ref{alg:Recursion-radar_measurement}.
We compute the estimate at each time step by using the posterior mean
of the hypothesis with highest weight in the $\delta$-GLMB filter.
Due to the IID cluster birth process, there is track switching at
all time steps for the trajectories born at the same time in the $\delta$-GLMB
filter estimate, which is shown in Figure \ref{fig:Estimates_simulation_deltaGLMB}.
This was previously indicated in Example \ref{exa:Label_mixing} and
is not a desirable situation that can be avoided by using the joint
density over the sequence of labelled sets, though there is no closed-form
expression in the literature, or by using the set of trajectories
as state variable, as we propose in this paper. 

\begin{figure}
\begin{centering}
\subfloat[]{\begin{centering}
\includegraphics[scale=0.29]{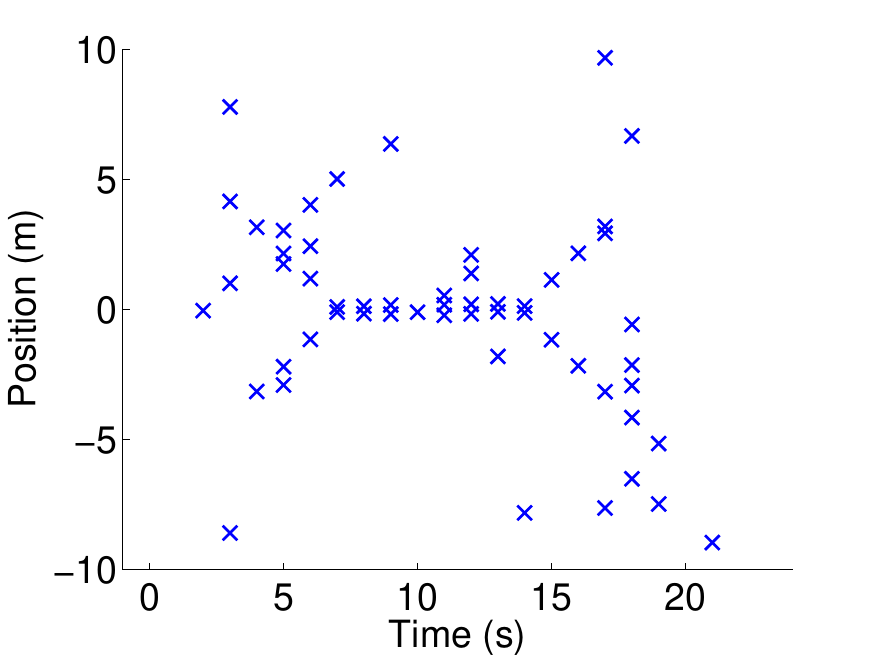}
\par\end{centering}

}\subfloat[0.21]{\begin{centering}
\includegraphics[scale=0.29]{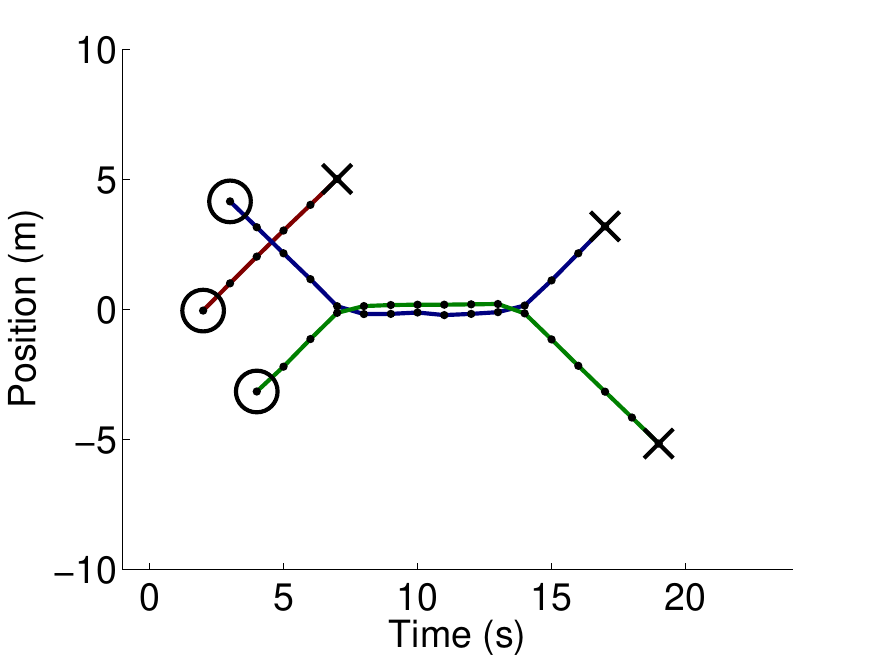}
\par\end{centering}

}
\par\end{centering}
\begin{centering}
\subfloat[0.14]{\begin{centering}
\includegraphics[scale=0.29]{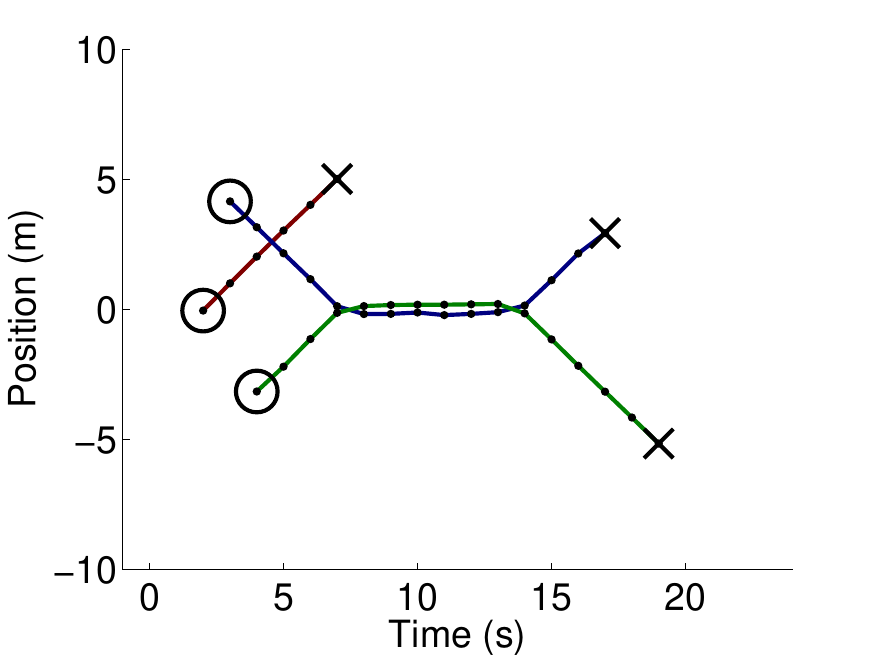}
\par\end{centering}

}\subfloat[0.08]{\begin{centering}
\includegraphics[scale=0.29]{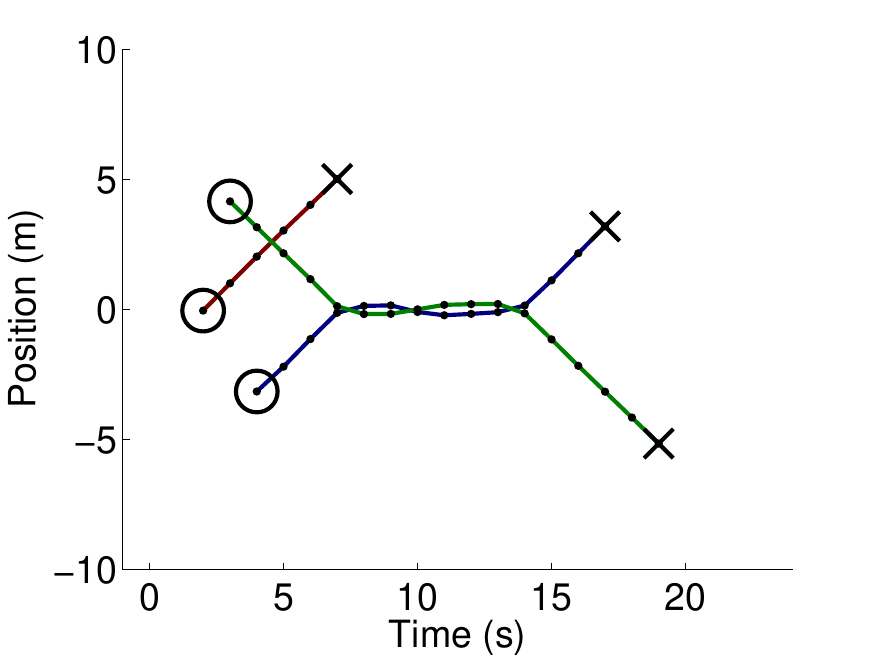}
\par\end{centering}

}
\par\end{centering}
\caption{\label{fig:Measurements-hypotheses_scenario3}Measurements and three
most likely global hypotheses. Their weights are given in the subfigure
captions. The third most likely global hypothesis includes a trajectory
switching.}

\end{figure}

We now consider another scenario with the same parameters but with
$\kappa\left(z\right)=\frac{1}{20}1_{\left(-10,10\right)}\left(z\right)$
and $R=10^{-3}$, i.e., lower clutter and higher measurement noise.
We show the observed measurements and the three most likely hypotheses
in Figure \ref{fig:Measurements-hypotheses_scenario3}. None of the
three most likely global hypotheses has two trajectories that start
at time 2 as the trajectories that were used to generate the measurements,
see Figure \ref{fig:Observed-measurements}(b). This is due to the
fact that one of the trajectories has not been detected at time 2,
see Figure \ref{fig:Measurements-hypotheses_scenario3}(a). The third
most likely global hypothesis includes a trajectory switching when
targets are in close proximity.

\subsection{Two dimensional scenario\label{subsec:Implementation-using-Murty's}}

We consider that a target state $x\in\mathbb{R}^{4}$ that includes
the position and velocity in a two dimensional space. The birth process
has parameters: $b=2$, $\beta_{1}\left(x\right)=\beta_{2}\left(x\right)=\mathcal{N}\left(x;\left[500,2,500,2\right]^{T},\mathrm{diag}\left(250^{2},5^{2},250^{2},5^{2}\right)\right)$,
$w_{B}\left(\oslash\right)=0.85$, $w_{B}\left(\left\{ 1\right\} \right)=w_{B}\left(\left\{ 2\right\} \right)=0.05$,
$w_{B}\left(\left\{ 1,2\right\} \right)=0.05$, where it should be
noted that the two targets can be born in a large area, not at a very
specific location. The single-target dynamic process parameters are:
$p_{S}=0.99$, $g\left(x^{i}|x^{i-1}\right)=\mathcal{N}\left(x^{i};Fx^{i-1},Q\right)$
with 
\begin{align*}
F=I_{2}\otimes & \left(\begin{array}{cc}
1 & \tau\\
0 & 1
\end{array}\right),\quad Q=qI_{2}\otimes\left(\begin{array}{cc}
\tau^{3}/3 & \tau^{2}/2\\
\tau^{2}/2 & \tau
\end{array}\right),
\end{align*}
where $\otimes$ denotes Kronecker product, $\tau$ is the sampling
time and $q$ is a parameter of the model. We use $\tau=1$, $q=0.25$
and we consider a total number of $k_{f}=150$ steps in the simulations. 

The intensity function of the Poisson clutter is $\kappa\left(z\right)=70\cdot\frac{1}{10^{6}}1_{A}\left(z\right)$,
which means that clutter is uniformly distributed in an area $\left(0,1000\right)\times\left(0,1000\right)$
and there is an average of $70$ clutter measurement per scan. The
position of each target is measured with parameters: $p_{D}=0.9$
and $l\left(z|x\right)=\mathcal{N}\left(z;Hx,R\right)$ where $R=I_{2}$
and
\begin{align*}
H= & \left(\begin{array}{cccc}
1 & 0 & 0 & 0\\
0 & 0 & 1 & 0
\end{array}\right).
\end{align*}

The prediction and update steps of the trajectory $\mathrm{MBM}_{01}$
filter are computed as indicated in Appendix \ref{sec:Appendix_Gaussian_implementation}.
The maximum number of global hypotheses we consider is $N_{h}=300$.
At each update step, we use ellipsoidal gating \cite{Kurien_inbook90}
with threshold 10 to consider only the relevant measurements for each
single trajectory hypothesis. In addition, for each global hypothesis,
we use Murty's algorithm to select the $k_{m}$-best new global hypotheses
without having to evaluate the weights of all the new global hypotheses.
The cost matrix of Murty's algorithm is the same as in the $\delta$-GLMB
filter, as the new data associations only depend on the target states
at the current time. As in \cite{Vo14}, for a global hypothesis with
a weight $w^{k|k-1}\left(\cdot\right)$, we set $k_{m}$ to $\left\lceil N_{h}\cdot w^{k|k-1}\left(\cdot\right)\right\rceil $.
Once the new global hypotheses are formed, pruning is performed to
keep the $N_{h}$ global hypotheses with highest weights. For the
prediction step, we use the $k_{p}$-shortest path algorithm to prune
the number of predicted hypotheses of the surviving trajectories,
as in \cite{Vo13}. We select $k_{p}$ as $\left\lceil 3\cdot N_{h}\cdot w^{k|k}\left(\cdot\right)\right\rceil $. 

Also, each trajectory density is propagated using an $L$-scan sliding
window. That is, as explained in \cite{Angel18_c}, an $L$-scan trajectory
density considers a joint density over the last $L$ time steps of
the trajectory and the rest of the time steps are considered independent.
Therefore, for past states beyond the $L$-scan window, we only need
to store the means and covariance matrices for the corresponding target
state at each each time step. Moreover, if one is only interested
in estimation, and not in the underlying uncertainty, one can simply
store the trajectory means outside the $L$-scan window for each trajectory
density. At each time step, we can estimate the true sets of trajectories,
for example, by considering the posterior means of the trajectories
for the global hypothesis with highest weight. 

\begin{figure}
\begin{centering}
\includegraphics[scale=0.5]{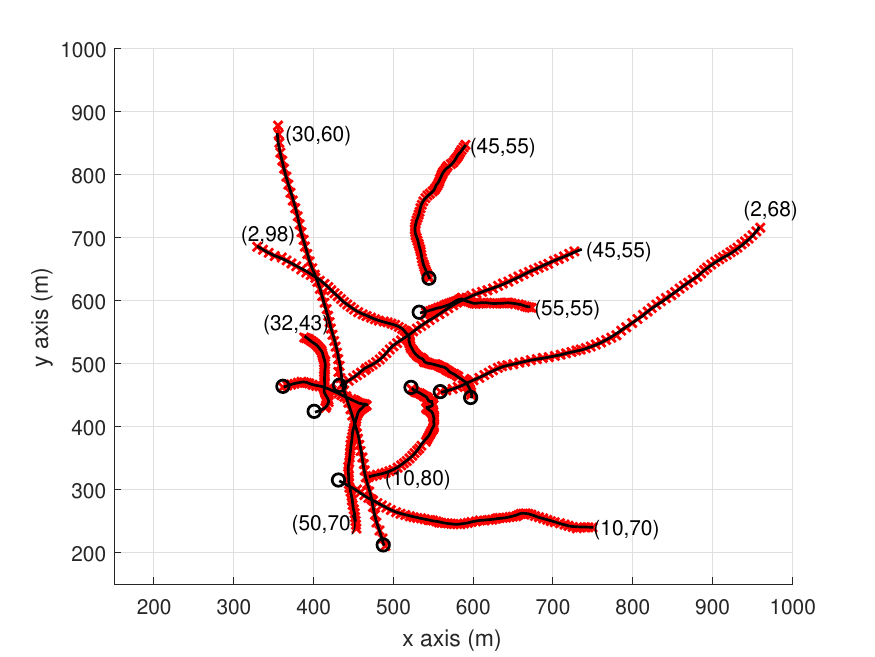}
\par\end{centering}
\caption{\label{fig:Murty_implementation}The ten true trajectories are shown
as black lines. The initial position of a true trajectory is marked
with a black circle and its $\left(t,i\right)$, starting time and
duration, is indicated next to its end position. The posterior means
of the trajectories for the global hypothesis with highest weight
of the trajectory $\mathrm{MBM}_{01}$ filter at the final time step
are shown with red lines and crosses. }
\end{figure}

We consider 10 true trajectories, which have been obtained by sampling
from the dynamic model, and are shown in Figure \ref{fig:Murty_implementation}.
We first consider one realisation of the measurements so that we can
also plot the posterior means of the trajectories for the global hypothesis
with highest weight at the end of the simulation. In Figure \ref{fig:Murty_implementation},
the single trajectory densities have been propagated using $L=2$.
We can see that the conjugate prior can be used to successfully estimate
the true set of trajectories. No false trajectories are reported,
though some trajectories are detected after some delay. It should
be noted that changing $L$ does not affect the weights of the hypotheses
or the available information on the targets at the current time step.
Increasing $L$ only improves the knowledge we have about past states
of a trajectory. Also, if we set $L=1$, we have an algorithm that
does not perform trajectory smoothing. That is, for $L=1$, we run
a Kalman filter to obtain the mean and covariance matrix at the current
time, for each single trajectory hypothesis, and all the previously
obtained means (and optionally the covariance matrices) are stored.
In fact, for $L=1$ and the pruning of the predicted and updated hypotheses
explained above, the trajectory $\mathrm{MBM}_{01}$ filter and the
$\delta$-GLMB filter perform the same computational operations. However,
the trajectory $\mathrm{MBM}_{01}$ filter with $L=1$ requires more
storage than $\delta$-GLMB to keep information about past states
of the trajectories for all single trajectory hypotheses, including
dead ones. Also, estimation is performed in a different manner in
which a trajectory estimate is obtained from the same single trajectory
hypothesis at all time steps. This avoids track switching of the type
shown in Figure \ref{fig:Equally-likely-trajectories}(b). 

The running times to process the 150 time steps using a non-optimised
Matlab implementation on an Intel Xeon CPU at 3.5 GHz for different
values of $L$ are: 143.2 s (1-scan/$\delta$-GLMB filter), 144.4
s (2-scan), 145.7 s (5-scan) and 159.7 s (10-scan). It is important
to notice that the computational burden for $L$ between 1 and 5 is
quite similar. This implies that, for values of $L$ between 1 and
5, most of the computational burden is related to how hypotheses are
handled, and improving the knowledge we have about past trajectory
states requires a small computational cost.

We would like to mention that, with the considered pruning parameters,
the global hypothesis with highest weight does not consider two targets
born at the same time, despite the fact that this is the ground truth
situation. If we increase the number of global hypotheses to 4000,
the global hypothesis with highest weight includes two targets born
at time step 2. In this case, the $\delta$-GLMB filter shows track
switching for the two targets born at time step 2, as was illustrated
in Figure \ref{fig:Estimates_simulation_deltaGLMB}. This illustration
is omitted in this example as it has been illustrated before. This
track switching can also be avoided by considering sequences of sets
of labelled targets. However, the only algorithm available in the
literature \cite{Vu14} is a batch algorithm, so it is not addressing
the same problem, and is based on sampling of trajectories and data
associations, which is not necessary for linear/Gaussian problems.

We proceed to analyse the error in the final trajectory estimates
obtained by the trajectory $\mathrm{MBM}_{01}$ filter and the $\delta$-GLMB
filter using Monte Carlo simulation with $N_{mc}=100$ runs. In order
to do so, we calculate the root mean square optimal assignment sub-pattern
assignment (OSPA) \cite{Schuhmacher08_b,Schuhmacher08} error and
the root mean square GOSPA \cite{Rahmathullah17} for the position
elements across all time steps. Both metrics use the Euclidean metric
as the base metric, $p=2$ and $c=15$. The GOSPA metric $d\left(\cdot,\cdot\right)$
uses $\alpha=2$ as this choice enables the decomposition of the error
into localisation error for properly detected targets and costs for
missed and false targets \cite{Rahmathullah17} such that
\begin{align*}
d^{2}\left(\mathbf{x}^{k},\mathbf{\hat{x}}_{i}^{k}\right) & =c_{l}^{2}\left(\mathbf{x}^{k},\mathbf{\hat{x}}_{i}^{k}\right)+c_{m}^{2}\left(\mathbf{x}^{k},\mathbf{\hat{x}}_{i}^{k}\right)+c_{f}^{2}\left(\mathbf{x}^{k},\mathbf{\hat{x}}_{i}^{k}\right)
\end{align*}
where $c_{l}^{2}\left(\cdot,\cdot\right)$, $c_{m}^{2}\left(\cdot,\cdot\right)$,
and $c_{f}^{2}\left(\cdot,\cdot\right)$ are the squared cost for
localisation error, missed targets and false targets, respectively. 

The root mean square GOSPA/OSPA error across all time steps is
\begin{align}
\mathrm{RMSGOSPA} & =\sqrt{\frac{1}{N_{mc}k_{f}}\sum_{i=1}^{N_{mc}}\sum_{k=1}^{k_{f}}d^{2}\left(\mathbf{x}^{k},\mathbf{\hat{x}}_{i}^{k}\right)}\label{eq:RMS_GOSPA}
\end{align}
where $\mathbf{x}^{k}$ is the set of targets at time $k$ and $\mathbf{\hat{x}}_{i}^{k}$
is the estimated set of targets at time $k$ in the $i$-th Monte
Carlo runs. The resulting errors as well as the root mean square errors
for the decomposition of the GOSPA metric are shown in Table \ref{tab:GOSPA_errors}.
Increasing the value of $L$ lowers GOSPA/OSPA errors. From the GOSPA
decomposition, we can see that increasing $L$ lowers the localisation
error for properly detected targets, but does not change the cost
for missed and false targets. The cost for missed and false targets
does not change, as in this case, $L=1$ is sufficient to estimate
the target states within a radius of $c$. In addition, the main source
of GOSPA error for any $L$ are missed target costs and localisation
errors. False target costs are smaller. For any value of $L$, the
proposed algorithm based on sets of trajectories outperforms the $\delta$-GLMB
filter for both GOSPA and OSPA metrics.

In Table \ref{tab:GOSPA_errors}, we also show the root mean square
error for the position elements using the metric for sets of trajectories
in \cite{Rahmathullah16_prov2} based on linear programming (LP),
with parameters $p=2$ and $c=15$ and $\gamma=1$. This metric, apart
from penalising localisation errors for properly detected targets,
missed and false targets, it also penalises track switches, and it
can be decomposed into its different components in a manner similar
to GOSPA. As expected, the $\delta$-GLMB filter has a higher error
related to track switches than the trajectory $\mathrm{MBM}_{01}$
filter. 

\begin{table}
\caption{\label{tab:GOSPA_errors}Root mean square OSPA/GOSPA/LP metric errors}
\begin{centering}
\par\end{centering}
\begin{centering}
\begin{tabular}{c|cccc}
\hline 
Algorithm &
\multicolumn{3}{c}{Trajectory $\mathrm{MBM}_{01}$ } &
$\delta$-GLMB\tabularnewline
\hline 
$L$ &
1 &
2 &
5 &
-\tabularnewline
\hline 
OSPA &
38.54 &
37.28 &
36.93 &
64.83\tabularnewline
\hline 
GOSPA &
58.34 &
53.37 &
51.99 &
90.56\tabularnewline
GOSPA-Localisation &
33.22 &
23.40 &
20.06 &
31.52\tabularnewline
GOSPA-Missed &
44.18 &
44.18 &
44.18 &
74.73\tabularnewline
GOSPA-False &
18.67 &
18.67 &
18.67 &
40.29\tabularnewline
\hline 
LP trajectory metric &
58.35 &
53.37 &
51.99 &
91.12\tabularnewline
LP-Localisation &
33.21 &
23.40 &
20.06 &
31.52\tabularnewline
LP-Missed &
44.18 &
44.18 &
44.18 &
74.73\tabularnewline
LP-False &
18.67 &
18.67 &
18.67 &
40.51\tabularnewline
LP-Switches &
0.66 &
0.63 &
0.63 &
2.32\tabularnewline
\hline 
\end{tabular}
\par\end{centering}
\centering{}
\end{table}

\section{Conclusions\label{sec:Conclusions}}

In this paper, we have proposed the set of trajectories as the variable
of interest in MTT and have indicated how to use Mahler's RFS framework
on this variable to fully characterise the MTT problem with targets
without a priori identification. Using the multitrajectory density
on sets of trajectories, we can answer all trajectory related questions.
Set of trajectories also enable us to define metrics with physical
interpretation, which can be used to obtain optimal estimators and
evaluate algorithms. Therefore, sets of trajectories along with Mahler's
RFS framework enable us to perform all the tasks of MTT.

We have also derived a conjugate multitrajectory density that gives
rise to the trajectory $\mathrm{MBM}_{01}$ filter to compute multitrajectory
filtering density.  In addition, we have described relations with
other labelled and unlabelled RFS models and MHT. Finally, we have
illustrated via simulations how we can perform MTT using sets of trajectories. 

Future work includes the specification of the measure theory aspects
for sets of trajectories, the development of computationally efficient
algorithms for MTT based on sets of trajectories and the inclusion
of target spawning.

\appendices{}

\section{\label{sec:Appendix_FISST}}

FISST can be applied with a single object space that is locally compact,
Hausdorff and second-countable (LCHS) \cite[Sec. 2.2.2]{Mahler_book14}.
In Appendix \ref{subsec:Appendix_LCHS}, we prove that the trajectory
space $T_{\left(k'\right)}$ up to time step $k'$ is LCHS so we can
use FISST with sets of trajectories. In Appendix \ref{subsec:Appendix_Integrals},
we explain how the single trajectory integral and the set integral
are defined. 

\subsection{\textmd{\normalsize{}The space $T_{\left(k'\right)}$ is LCHS\label{subsec:Appendix_LCHS}}}

The trajectory space $T_{\left(k'\right)}$ up to a certain time step
$k'$ is the disjoint union, which is also called the direct sum \cite{Engelking_book89},
of a finite number of subspaces, for which one can define the disjoint
union topology \cite{Willard_book70}. As $D^{i}$ is a Euclidean
space, $D^{i}$ is locally compact and Hausdorff. Then, it is direct
to show that $\left\{ t\right\} \times D^{i}$ is locally compact
and Hausdorff. The disjoint union of Hausdorff spaces is Hausdorff
\cite{Crossley_book05}, so $T_{\left(k'\right)}$ is Hausdorff. The
disjoint union of locally compact spaces is locally compact \cite{Engelking_book89},
so $T_{\left(k'\right)}$ is locally compact. 

In order to prove that $T_{\left(k'\right)}$ is second-countable,
which is synonymous to completely separable and implies separability
\cite{Simmons_book63}, we first recall two definitions. A space $E$
is second-countable if and only if its topology has a countable base
\cite[page 108]{Willard_book70}. A base in a space $E$ is a collection
of open sets in $E$ such that every open set in $E$ can be written
as a union of elements of the base \cite[page 38]{Willard_book70}.
The space $D^{i}=\mathbb{R}^{in_{x}}$, with the usual topology, is
second-countable with a countable base $\mathcal{B}_{1}\left(i\right)$,
which denotes the collection of open balls centred on all rational
points in $D^{i}$ with all rational radius \cite[page 100]{Simmons_book63}. 

We proceed to prove that $T_{\left(k'\right)}$ is second-countable
with countable base $\mathcal{B}_{2}=\uplus_{t\text{=1}}^{k'}\left\{ t\right\} \times\uplus_{i=1}^{k'-t+1}\mathcal{B}_{1}\left(i\right)$.
It is direct to see that the space $\left\{ t\right\} \times\mathbb{R}^{in_{x}}$
is second-countable with a countable base $\left\{ t\right\} \times\mathcal{B}_{1}\left(i\right)$.
In addition, due to the disjoint union topology, a set in $S\subseteq T_{\left(k'\right)}$
is open if and only if $S\cap\left(\left\{ t\right\} \times D^{i}\right)$
is open for all $\left(t,i\right)\in I_{(k')}$. Also, since $S$
is the union of the sets $S\cap\left(\left\{ t\right\} \times D^{i}\right)$
for all $\left(t,i\right)\in I_{(k')}$, it follows that any open
set $S$ can be written as a union of elements of $\mathcal{B}_{2}$,
which is a countable set. This proves that the single trajectory space
is second-countable.

\subsection{Integrals\label{subsec:Appendix_Integrals}}

In FISST, if the single object space is the disjoint union of subspaces,
each endowed with an integral, the single object integral is the sum
of the integrals over each subspace \cite[Sec. 3.5.3]{Mahler_book14}.
In the trajectory case, $T_{\left(k'\right)}=\uplus_{\left(t,i\right)\in I_{(k')}}\left\{ t\right\} \times D^{i}$
where $\left\{ t\right\} \times D^{i}$ is the Cartesian product of
a single element set and a Euclidean space, for which the discrete-Lebesgue
integral exists. Therefore, the single trajectory integral is given
by (\ref{eq:single_trajectory_integral}). Following \cite[Sec. 3.5.3]{Mahler_book14},
the corresponding set integral on $\mathcal{F}\left(T_{\left(k'\right)}\right)$
is given by (\ref{eq:set_integral_trajectory}). Single object spaces
that are the disjoint union of spaces of different dimensionality,
as the single trajectory space, have been used for RFSs in \cite[Sec. 2.2.2, Sec. 11.6, Chap. 18]{Mahler_book14}\cite{Mahler10}. 

Based on set integrals and multi-object densities, one can define
probability generating functionals \cite[Eq. (44)]{Mahler03}, which
can be used to derive filters. It is also possible to define measure
theoretic integrals for functions on sets in LCHS spaces \cite[App. B]{Vo05},
which can be used to define probability densities, though it is beyond
the scope of this paper to present the details.

\section{\label{sec:Appendix_Integration_janossy}}

We explain the connection between multitarget densities and Janossy
densities \cite{Daley_book03} and how to calculate probabilities.

\subsection{Target case}

A multitarget density packages the entire family of Janossy densities
into a single function \cite[p. 1161]{Mahler03}. Given a multiobject
density $\pi\left(\cdot\right)$, we can obtain the equivalent $n$th
Janossy density $j_{n}\left(\cdot\right)$ as \cite[Sec. II.D]{Mahler03},
\begin{align}
j_{n}\left(x_{1},...,x_{n}\right) & =\pi\left(\left\{ x_{1},...,x_{n}\right\} \right).\label{eq:janossy_FISST_equivalence}
\end{align}
The Janossy density $j_{n}\left(\cdot\right)$ is the density of the
$n$th Janossy measure w.r.t. the Lebesgue measure on $D^{n}$ \cite[p. 124]{Daley_book03}.
Then,
\begin{align}
j_{n}\left(x_{1},...,x_{n}\right)dx_{1}...dx_{n}\label{eq:janossy_integration}
\end{align}
is the probability that there are exactly $n$ points in the process,
one in each of the $n$ distinct infinitesimal regions $\left(x_{i},x_{i}+dx_{i}\right)\subset D$
\cite{Daley_book03}.

We define a region $A=\uplus_{n\text{=0}}^{\infty}A_{n}$ where $A_{n}\subseteq\mathcal{F}\left(D\right)$
is a set that contains sets with $n$ elements in $D$. Then, for
$\mathbf{x}$ distributed as $\pi\left(\cdot\right)$, 
\begin{align}
P\left(\mathbf{x}\in A\right) & =\sum_{n=0}^{\infty}P\left(\left\{ x_{1},...,x_{n}\right\} \in A_{n}\right)\nonumber \\
 & =\sum_{n=0}^{\infty}P\left(x_{1:n}\in\chi^{-1}\left(A_{n}\right)\right)\nonumber \\
 & =\sum_{n=0}^{\infty}\frac{1}{n!}\int_{\chi^{-1}\left(A_{n}\right)}j_{n}\left(x_{1:n}\right)dx_{1:n}\nonumber \\
 & =\sum_{n=0}^{\infty}\frac{1}{n!}\int_{\chi^{-1}\left(A_{n}\right)}\pi\left(\left\{ x_{1},...,x_{n}\right\} \right)dx_{1:n}.\label{eq:target_janossy_integration}
\end{align}
where $\chi\left(\cdot\right)$ denotes the mapping from vectors to
sets, as indicated in Section \ref{subsec:Probability-and-integration}
for trajectories or in \cite{Vo05}.

\subsection{Trajectory case}

By analogy to (\ref{eq:janossy_FISST_equivalence})-(\ref{eq:janossy_integration}),
for a multitrajectory density $\pi\left(\cdot\right)$, 
\begin{align*}
\pi\left(\left\{ \left(t_{1},x_{1}^{1:i_{1}}\right),...,\left(t_{n},x_{n}^{1:i_{n}}\right)\right\} \right)dx_{1}^{1:i_{1}}...dx_{n}^{1:i_{n}}
\end{align*}
is the probability that there are exactly $n$ distinct trajectories,
with starting times and lengths $t_{1},i_{1},...,t_{n},i_{n}$ in
infinitesimal regions $\left(x_{j}^{1:i_{j}},x_{j}^{1:i_{j}}+dx_{j}^{1:i_{j}}\right)\subset D^{i_{j}}$.
For $\mathbf{X}$ distributed according to $\pi\left(\cdot\right)$
and a region $A=\uplus_{n\text{=0}}^{\infty}A_{n}$, where $A_{n}\subseteq\mathcal{F}\left(T_{\left(k'\right)}\right)$
is a set that contains sets with $n$ elements in $T_{\left(k'\right)}$,
we have
\begin{align*}
P\left(\mathbf{X}\in A\right) & =\sum_{n=0}^{\infty}P\left(\left\{ X_{1},...,X_{n}\right\} \in A_{n}\right)\\
 & =\sum_{n=0}^{\infty}\frac{1}{n!}\int_{\chi^{-1}\left(A_{n}\right)}\pi\left(\left\{ X_{1},...,X_{n}\right\} \right)dX_{1:n}.
\end{align*}

\section{\label{sec:Appendix_general_prediction}}

In this appendix, we prove Theorem \ref{thm:General_prediction}.
First, we provide the transition density. Given $X=\left(t,x^{1:i}\right)$
with $t+i-1<k$ and $Y=\left(t',y^{1:i'}\right)$, the single trajectory
transition density at time $k$ is
\begin{align}
g^{k}\left(Y\left|X\right.\right) & =\left|\tau^{k-1}\left(X\right)\right|\left[\vphantom{\delta\left(\left(t',y^{1:i'-1}\right)-X\right)}\left(1-p_{S}\left(x^{i}\right)\right)\delta_{X}\left(Y\right)\right.\nonumber \\
 & \quad\left.+p_{S}\left(x^{i}\right)g\left(y^{i'}|x^{i}\right)\delta_{X}\left(\left(t',y^{1:i'-1}\right)\right)\right]\nonumber \\
 & \quad+\left(1-\left|\tau^{k-1}\left(X\right)\right|\right)\delta_{X}\left(Y\right).\label{eq:single_trajectory_transition}
\end{align}
That is, if the trajectory has died before time step $k-1$, the trajectory
remains unaltered with probability one. If the trajectory exists at
time step $k-1$, it remains unaltered (target dies) with probability
$\left(1-p_{S}\left(\cdot\right)\right)$ or the last target state
is generated according to the single target transition density with
probability $p_{S}\left(\cdot\right)$. Without target births, the
number of trajectories remains unchanged so the transition density
for the sets of trajectories is \cite[Eq. (12.91)]{Mahler_book07}
\begin{align}
g^{k}\left(\left\{ Y_{1},...,Y_{n}\right\} \left|\left\{ X_{1},...,X_{n}\right\} \right.\right) & =\sum_{\sigma\in\Gamma_{n}}\prod_{j=1}^{n}g^{k}\left(Y_{\sigma_{j}}\left|X_{j}\right.\right)\label{eq:multitrajectory_transition_no_birth}
\end{align}
where $\Gamma_{n}$ is the set that includes all the permutations
of $\left(1,...,n\right)$.

We evaluate $\pi^{k|k-1}\left(\cdot\right)$ for $\mathbf{W}\uplus\mathbf{X}\uplus\mathbf{Y}\uplus\mathbf{Z}$,
where these sets are defined in Theorem \ref{thm:General_prediction}.
The multitrajectory density for set $\mathbf{X}\uplus\mathbf{Y}\uplus\mathbf{Z}$
(no births) is denoted by $\pi_{S}^{k|k-1}\left(\cdot\right)$ and
the multitrajectory density for set $\mathbf{W}$ by $\beta^{k}\left(\cdot\right)$.
Using the convolution formula \cite[Sec. 4.2.3]{Mahler_book14}, we
have 
\begin{align*}
 & \pi^{k|k-1}\left(\mathbf{W}\uplus\mathbf{X}\uplus\mathbf{Y}\uplus\mathbf{Z}\right)\\
 & =\sum_{\mathbf{A}\subseteq\mathbf{W}\uplus\mathbf{X}\uplus\mathbf{Y}\uplus\mathbf{Z}}\pi_{S}^{k|k-1}\left(\mathbf{W}\uplus\mathbf{X}\uplus\mathbf{Y}\uplus\mathbf{Z}\setminus\mathbf{A}\right)\beta^{k}\left(\mathbf{A}\right).
\end{align*}
As $\beta^{k}\left(\cdot\right)$ is zero unless evaluated at a set
of new born trajectories, $\beta^{k}\left(\mathbf{A}\right)$ is different
from zero only if $\mathbf{A}=\mathbf{W}$, which yields
\begin{align*}
\pi^{k|k-1}\left(\mathbf{W}\uplus\mathbf{X}\uplus\mathbf{Y}\uplus\mathbf{Z}\right) & =\pi_{S}^{k|k-1}\left(\mathbf{X}\uplus\mathbf{Y}\uplus\mathbf{Z}\right)\beta^{k}\left(\mathbf{W}\right)
\end{align*}
where
\begin{align*}
\beta^{k}\left(\mathbf{W}\right) & =\beta_{\tau}\left(\tau^{k}\left(\mathbf{W}\right)\right)
\end{align*}
as the states of new born trajectories are distributed according to
the multiobject density $\beta_{\tau}\left(\cdot\right)$ and the
birth time and length are deterministic. 

On the other hand, the multitrajectory density for the surviving targets
is given by the prediction step, with the transition density without
target births which is given in (\ref{eq:multitrajectory_transition_no_birth}).
Then,
\begin{align*}
\pi_{S}^{k|k-1}\left(\mathbf{X}\uplus\mathbf{Y}\uplus\mathbf{Z}\right) & =\int g^{k}\left(\mathbf{X}\uplus\mathbf{Y}\uplus\mathbf{Z}\left|\mathbf{W}'\right.\right)\pi^{k-1}\left(\mathbf{W}'\right)\delta\mathbf{W}',
\end{align*}
Partitioning $\mathbf{W}'=\mathbf{D}\uplus\mathbf{A}$, where $\mathbf{D}$
and $\mathbf{A}$ represent dead and alive trajectories at time $k-1$,
the set integral over $\mathbf{W}'$ can be calculated as the set
integral over $\mathbf{D}$ and $\mathbf{A}$ \cite[Sec. 3.5.3]{Mahler_book14}
\begin{align*}
 & \pi_{S}^{k|k-1}\left(\mathbf{X}\uplus\mathbf{Y}\uplus\mathbf{Z}\right)\\
 & \quad=\int\int g^{k}\left(\mathbf{X}\uplus\mathbf{Y}\uplus\mathbf{Z}\left|\mathbf{D}\uplus\mathbf{A}\right.\right)\pi^{k-1}\left(\mathbf{D}\uplus\mathbf{A}\right)\delta\mathbf{D}\delta\mathbf{A}\\
 & \quad=\int g^{k}\left(\mathbf{X}\uplus\mathbf{Y}\left|\mathbf{A}\right.\right)\pi^{k-1}\left(\mathbf{Z}\uplus\mathbf{A}\right)\delta\mathbf{A}.
\end{align*}
Evaluating this expression for $\mathbf{X}=\left\{ X_{1},...,X_{m}\right\} $
and $\mathbf{Y}=\left\{ Y_{1},...,Y_{n}\right\} $ and using (\ref{eq:multitrajectory_transition_no_birth}),
we have
\begin{align}
 & \pi_{S}^{k|k-1}\left(\mathbf{X}\uplus\mathbf{Y}\uplus\mathbf{Z}\right)\nonumber \\
 & =\frac{1}{\left(m+n\right)!}\int\sum_{\sigma\in\Gamma_{m+n}}\prod_{j=1}^{_{m}}g^{k}\left(X_{j}|A_{\sigma\left(j\right)}\right)\nonumber \\
 & \times\prod_{j=m+1}^{_{m+n}}g^{k}\left(Y_{j-m}|A_{\sigma\left(j\right)}\right)\nonumber \\
 & \times\pi^{k-1}\left(\mathbf{Z}\uplus\left\{ A_{1},...,A_{m+n}\right\} \right)dA_{1:m+n}\nonumber \\
 & =\int\prod_{j=1}^{_{m}}g^{k}\left(X_{j}|A_{j}\right)\prod_{j=m+1}^{_{m+n}}g^{k}\left(Y_{j-m}|A_{j}\right)\nonumber \\
 & \quad\quad\times\pi^{k-1}\left(\mathbf{Z}\uplus\left\{ A_{1},...,A_{m+n}\right\} \right)dA_{1:m+n}.\label{eq:pdf_surviving_append}
\end{align}
The proof is finished by substituting (\ref{eq:single_trajectory_transition})
into (\ref{eq:pdf_surviving_append}).

\section{\label{sec:Appendix_conjugate_prior}}

In this appendix, we prove that the birth model (\ref{eq:birth_model_radar})
is an $\mathrm{MBM}_{01}$. We also prove Lemmas \ref{lem:Prediction_radar}
and \ref{lem:Update_radar}. 

\subsection{Birth model}

We write (\ref{eq:birth_model_radar}) as
\begin{align*}
 & \beta_{\tau}\left(\left\{ x_{1},...,x_{n}\right\} \right)\\
 & \quad=\sum_{l_{1:n}}^{\neq}\left(\sum_{L\subseteq\mathbb{N}_{b}}w_{B}\left(L\right)\delta_{L}\left(\left\{ l_{1},...,l_{n}\right\} \right)\right)\prod_{j=1}^{n}\beta_{l_{j}}\left(x_{j}\right)\\
 & \quad=\sum_{L\subseteq\mathbb{N}_{b}}w_{B}\left(L\right)\left(\sum_{l_{1:n}}^{\neq}\delta_{L}\left(\left\{ l_{1},...,l_{n}\right\} \right)\prod_{j=1}^{n}\beta_{l_{j}}\left(x_{j}\right)\right)
\end{align*}
where $\delta_{L}\left(\cdot\right)$ is the Kronecker delta for sets:
$\delta_{L}\left(\left\{ l_{1},...,l_{n}\right\} \right)=1$ if $\left\{ l_{1},...,l_{n}\right\} =L$
and zero otherwise. From \cite[Eq. (11.133)]{Mahler_book07}, we can
see that the factor inside the bracket is a multi-Bernoulli density
such that the probabilities of existence of the Bernoulli components
whose indices belong to $L$ are one and the rest are zero. The corresponding
single-target density for $l_{j}\in\mathbb{N}_{b}$ is $\beta_{l_{j}}\left(\cdot\right)$.
Therefore, the birth model is a mixture of multi-Bernoulli densities
whose existence probabilities are either 0 or 1 and the weight of
each mixture component is given by $w_{B}\left(L\right)$ $L\subseteq\mathbb{N}_{b}$.
Analogously, one can check that the multitrajectory filtering and
predicted densities, which are given by (\ref{eq:posterior_radar})
and (\ref{eq:prior_radar}), are also $\mathrm{MBM}_{01}$. 

\subsection{Prediction}

We consider the sets $\mathbf{W}=\left\{ W_{1},...,W_{n_{w}}\right\} $,
$\mathbf{X}=\left\{ X_{1},...,X_{n_{x}}\right\} $, $\mathbf{Y}=\left\{ Y_{1},...,Y_{n_{y}}\right\} $
and $\mathbf{Z}=\left\{ Z_{1},...,Z_{n_{z}}\right\} $, as defined
in Theorem \ref{thm:General_prediction}. We also denote $n_{xyz}=n_{x}+n_{y}+n_{z}$,
$n_{xyzw}=n_{x}+n_{y}+n_{z}+n_{w}$ and $X_{j}=\left(t_{j},x_{j}^{1:i_{j}}\right)$.
Substituting (\ref{eq:birth_model_radar}) and (\ref{eq:posterior_radar})
into Theorem \ref{thm:General_prediction} yields
\begin{align*}
 & \pi^{k|k-1}\left(\mathbf{W}\uplus\mathbf{X}\uplus\mathbf{Y}\uplus\mathbf{Z}\right)\\
 & =\sum_{h_{1:n_{xyz}}^{k-1|k-1}}^{\neq}w^{k-1|k-1}\left(\left\{ h_{1}^{k-1|k-1},...,h_{n_{xyz}}^{k-1|k-1}\right\} \right)\\
 & \times\prod_{j=1}^{n_{x}}\left[g\left(x_{j}^{i_{j}}\left|x_{j}^{i_{j}-1}\right.\right)p_{S}\left(x_{j}^{i_{j}-1}\right)\right.\\
 & \,\left.\times p^{k-1|k-1}\left(t_{j},x_{j}^{1:i_{j}-1}|h_{j}^{k-1|k-1}\right)\right]\\
 & \,\times\prod_{j=1}^{n_{y}}\left[\left(1-p_{S}\left(Y_{j}^{|}\right)\right)p^{k-1|k-1}\left(Y_{j}|h_{j+n_{x}}^{k-1|k-1}\right)\right]\\
 & \,\times\prod_{j=1}^{n_{z}}p^{k-1|k-1}\left(Z_{j}|h_{j+n_{x}+n_{y}}^{k-1|k-1}\right)\\
 & \,\times\sum_{h_{n_{xyz}+1:n_{xyzw}}^{k|k-1}}^{\neq}w_{B}^{k}\left(\left\{ h_{n_{xyz}+1}^{k|k-1},...,h_{n_{xyzw}}^{k|k-1}\right\} \right)\\
 & \,\times\prod_{j=1}^{n_{w}}p_{N}^{k|k-1}\left(W_{j}|h_{j+n_{xyz}}^{k|k-1}\right)
\end{align*}
where we should note that $w_{B}^{k}\left(\cdot\right)$ is zero for
hypotheses that are not contained in $\mathbb{N}^{k}$. 

We can also express the above summations in terms of the hypotheses
$h_{j}^{k|k-1}$ instead of $h_{j}^{k-1|k-1}$. The relation between
the hypotheses is
\begin{align*}
h_{j}^{k-1|k-1} & =\begin{cases}
\left(h_{j}^{k|k-1}\right)^{-} & j=1,...,n_{x}\\
h_{j}^{k|k-1} & j=n_{x}+1,...,n_{xyz}
\end{cases}
\end{align*}
where $\left(h_{j}^{k|k-1}\right)^{-}$ is defined in Lemma \ref{lem:Prediction_radar}.
We can then write

\begin{align*}
 & \pi^{k|k-1}\left(\mathbf{W}\uplus\mathbf{X}\uplus\mathbf{Y}\uplus\mathbf{Z}\right)\\
 & =\sum_{h_{1:n_{xyzw}}^{k|k-1}}^{\neq}w^{k-1|k-1}\left(\left\{ \left(h_{1}^{k|k-1}\right)^{-},...,\left(h_{n_{x}}^{k|k-1}\right)^{-}\right\} \right.\\
 & \quad\left.\vphantom{\left(h_{1}^{k|k-1}\right)^{-}}\cup\left\{ h_{n_{x}+1}^{k|k-1},...,h_{n_{x}+n_{y}}^{k|k-1}\right\} \cup\left\{ h_{n_{x}+n_{y}+1}^{k|k-1},...,h_{n_{xyz}}^{k|k-1}\right\} \right)\\
 & \,\times w_{B}^{k}\left(\left\{ h_{n_{xyz}+1}^{k|k-1},...,h_{n_{xyzw}}^{k|k-1}\right\} \right)\\
 & \,\times\prod_{j=1}^{n_{x}}\gamma_{S}\left(h_{j}^{k|k-1}\right)\prod_{j=1}^{n_{y}}\gamma_{D}\left(h_{j+n_{x}}^{k|k-1}\right)\prod_{j=1}^{n_{x}}p_{S}^{k|k-1}\left(X_{j}|h_{j}^{k|k-1}\right)\\
 & \,\times\prod_{j=1}^{n_{y}}p_{D}^{k|k-1}\left(Y_{j}|h_{j+n_{x}}^{k|k-1}\right)\prod_{j=1}^{n_{z}}p^{k-1|k-1}\left(Z_{j}|h_{j+n_{x}+n_{y}}^{k|k-1}\right)\\
 & \,\times\prod_{j=1}^{n_{w}}p_{N}^{k|k-1}\left(W_{j}|h_{j+n_{xyz}}^{k|k-1}\right).
\end{align*}
Note that, as $\mathbf{X}$, $\mathbf{Y}$, $\mathbf{Z}$ and $\mathbf{W}$
represent surviving, dying, dead and new born trajectories, one must
have that sets of hypotheses $\mathcal{A}=\left\{ h_{1}^{k|k-1},...,h_{n_{x}}^{k|k-1}\right\} $,
$\mathcal{B}=\left\{ h_{n_{x}+1}^{k|k-1},...,h_{n_{x}+n_{y}}^{k|k-1}\right\} $,
$\mathcal{D}=\left\{ h_{n_{x}+n_{y}+1}^{k|k-1},...,h_{n_{xyz}}^{k|k-1}\right\} $
and $\mathcal{C}=\left\{ h_{n_{xyz}+1}^{k|k-1},...,h_{n_{xyzw}}^{k|k-1}\right\} $
represent surviving, dying, dead and new born trajectory hypotheses.
Otherwise, the corresponding terms in the sum over all hypotheses
are equal to zero. Therefore, given $\mathcal{A}$, $\mathcal{B}$,
$\mathcal{D}$ and $\mathcal{C}$, we can identify the weight $w^{k|k-1}\left(\cdot\right)$
of the predicted global hypothesis $\mathcal{A}\uplus\mathcal{B}\uplus\mathcal{C}\uplus\mathcal{D}$,
as the factors in front of the densities in the previous expression.
The resulting weight and single trajectory densities can be written
as in Lemma \ref{lem:Prediction_radar}, which concludes its proof.

\subsection{Update}

The predicted density (\ref{eq:prior_radar}) evaluated at a set $\mathbf{X}=\left\{ X_{1},...,X_{n_{x}}\right\} $
of alive trajectories and a set $\mathbf{Y}=\left\{ Y_{1},...,Y_{n_{y}}\right\} $
of dead trajectories can be written as 
\begin{align*}
 & \pi^{k|k-1}\left(\mathbf{X}\uplus\mathbf{Y}\right)\\
 & =\sum_{h_{1:n_{x}+n_{y}}^{k|k-1}}^{\neq}\prod_{j=1}^{n_{x}}p^{k|k-1}\left(X_{j}|h_{j}^{k|k-1}\right)\prod_{j=1}^{n_{y}}p^{k|k-1}\left(Y_{j}|h_{j+n_{x}}^{k|k-1}\right)\\
 & \,\times w^{k|k-1}\left(\left\{ h_{1}^{k|k-1},...,h_{n_{_{x}}}^{k|k-1}\right\} \cup\left\{ h_{n_{x}+1}^{k|k-1},...,h_{n_{x}+n_{_{y}}}^{k|k-1}\right\} \right)\\
 & =\sum_{h_{1:n_{x}}^{k|k-1}}^{\neq}\sum_{h_{n_{x}+1:n_{x}+n_{y}}^{k|k-1}}^{\neq}\prod_{j=1}^{n_{x}}p^{k|k-1}\left(X_{j}|h_{j}^{k|k-1}\right)\\
 & \,\times\prod_{j=1}^{n_{y}}p^{k|k-1}\left(Y_{j}|h_{j+n_{x}}^{k|k-1}\right)\\
 & \,\times w^{k|k-1}\left(\left\{ h_{1}^{k|k-1},...,h_{n_{_{x}}}^{k|k-1}\right\} \cup\left\{ h_{n_{x}+1}^{k|k-1},...,h_{n_{x}+n_{_{y}}}^{k|k-1}\right\} \right).
\end{align*}
As $\mathbf{X}$ and $\mathbf{Y}$ are sets of alive and dead trajectories,
the terms in the above summation are zero unless $h_{1:n_{x}}^{k|k-1}\in\left(\mathbb{U}^{k}\right)^{n_{x}}$
and $h_{n_{x}+1:n_{x}+n_{y}}^{k|k-1}\in\left(\mathbb{D}^{1:k}\right)^{n_{y}}$,
which represent alive and dead single trajectory hypotheses, respectively.

Substituting the previous equation and (\ref{eq:likelihood_radar})
into (\ref{eq:update_trajectories}), the updated density is 
\begin{align}
 & \pi^{k}\left(\mathbf{X}\uplus\mathbf{Y}\right)\nonumber \\
 & \propto\sum_{\theta\in\Theta_{n_{x},m}}\sum_{h_{1:n_{x}}^{k|k-1}}^{\neq}\sum_{h_{n_{x}+1:n_{x}+n_{y}}^{k|k-1}}^{\neq}\prod_{j=1}^{n_{y}}p^{k|k-1}\left(Y_{j}|h_{j+n_{x}}^{k|k-1}\right)\nonumber \\
 & \quad\times\prod_{j=1}^{n_{x}}\left[\psi_{\mathbf{z}^{k}}\left(X_{j}|\theta_{j}\right)p^{k|k-1}\left(X_{j}|h_{j}^{k|k-1}\right)\right]\nonumber \\
 & \,\times w^{k|k-1}\left(\left\{ h_{1}^{k|k-1},...,h_{n_{_{x}}}^{k|k-1}\right\} \cup\left\{ h_{n_{x}+1}^{k|k-1},...,h_{n_{x}+n_{_{y}}}^{k|k-1}\right\} \right)\nonumber \\
\nonumber \\
 & =\sum_{\left(h_{1}^{k|k-1},\theta_{1}\right)...\left(h_{n_{x}}^{k|k-1},\theta_{n_{x}}\right)}^{\neq}\sum_{h_{n_{x}+1:n_{x}+n_{y}}^{k|k-1}}^{\neq}\prod_{j=1}^{n_{y}}p^{k|k-1}\left(Y_{j}|h_{j+n_{x}}^{k|k-1}\right)\nonumber \\
 & \,\times\left[\prod_{j=1}^{n_{x}}\eta_{\mathbf{z}^{k}}\left(h_{j}^{k|k-1},\theta_{j}\right)\right]\left[\prod_{j=1}^{n_{x}}p_{U}^{k|k}\left(X_{j}|h_{j}^{k|k-1},\theta_{j}\right)\right]\nonumber \\
 & \,\times w^{k|k-1}\left(\left\{ h_{1}^{k|k-1},...,h_{n_{_{x}}}^{k|k-1}\right\} \cup\left\{ h_{n_{x}+1}^{k|k-1},...,h_{n_{x}+n_{_{y}}}^{k|k-1}\right\} \right).\label{eq:update_conjugate_prior_proof}
\end{align}
The updated weight $w^{k|k}\left(\cdot\right)$ of the new global
hypothesis $\left\{ \left(h_{1}^{k|k-1},\theta_{1}\right),...,.\left(h_{n_{x}}^{k|k-1},\theta_{n_{x}}\right)\right\} \cup\left\{ h_{n_{x}+1}^{k|k-1},...,h_{n_{x}+n_{_{y}}}^{k|k-1}\right\} $
is then proportional to
\begin{align*}
 & w^{k|k-1}\left(\left\{ h_{1}^{k|k-1},...,h_{n_{x}}^{k|k-1}\right\} \cup\left\{ h_{n_{x}+1}^{k|k-1},...,h_{n_{x}+n_{_{y}}}^{k|k-1}\right\} \right)\\
 & \,\times\prod_{j=1}^{n_{x}}\eta_{\mathbf{z}^{k}}\left(h_{j}^{k|k-1},\theta_{j}\right)
\end{align*}
where $\left\{ h_{1}^{k|k-1},...,h_{n_{x}}^{k|k-1}\right\} $ and
$\left\{ h_{n_{x}+1}^{k|k-1},...,h_{n_{x}+n_{_{y}}}^{k|k-1}\right\} $
represent sets of alive and dead trajectory hypotheses, respectively.
The updated densities are the ones indicated in (\ref{eq:update_conjugate_prior_proof}).
This concludes the proof of Lemma \ref{lem:Update_radar}. 

\section{\label{sec:Appendix_Gaussian_implementation}}

In this appendix, we provide more details of the computation of the
single trajectory densities of the trajectory $\mathrm{MBM}_{01}$
filter (Procedure \ref{alg:Recursion-radar_measurement}) with the
linear/Gaussian model used in Section \ref{sec:Illustrative-example}
with constant $p_{S}$ and $p_{D}$. The labelled trajectory $\mathrm{MBM}_{01}$
filter recursion would be analogous, as explained in Section \ref{subsec:Labelled-set-of-trajectories}.

We use the following notation to denote a Gaussian density on the
single trajectory space with a certain start time and length: 
\begin{align}
\mathcal{N}\left(\left(t,x^{1:i}\right);t^{k},m^{k},P^{k}\right) & \triangleq\mathcal{N}\left(x^{1:i};m^{k},P^{k}\right)\label{eq:Trajectory_Gaussian}
\end{align}
if $i=\mathrm{dim}\left(m^{k}\right)/n_{x}$ and $t=t^{k}$ or zero
otherwise. 

\subsection{Prediction}

We compute the quantities in Lemma \ref{lem:Prediction_radar}. Let
us denote the previous single trajectory densities for a given hypothesis
$h^{-}$ as
\begin{align}
p^{k|k}\left(X|h^{-}\right) & =\mathcal{N}\left(X;t_{h^{-}}^{k},m_{h^{-}}^{k},P_{h^{-}}^{k}\right).\label{eq:Trajectory_Gaussian_hypothesis}
\end{align}
Then, we have that $\gamma_{S}\left(h\right)=p_{S}$, and $w_{B}\left(\cdot\right)$
is provided in the problem formulation. Then,
\begin{align*}
p_{S}^{k+1|k}\left(t,x^{1:i}|h\right) & =\mathcal{N}\left(x^{i};Fx^{i-1},Q\right)\\
 & \quad\times\mathcal{N}\left(\left(t,x^{1:i-1}\right);t_{h^{-}}^{k},m_{h^{-}}^{k},P_{h^{-}}^{k}\right)\\
p_{D}^{k+1|k}\left(t,x^{1:i}|h\right) & =p^{k|k}\left(t,x^{1:i}|h\right)\\
p_{N}^{k+1|k}\left(t,x^{1}|l,k+1,1\right) & =\beta_{l}\left(x^{1}\right)\delta_{k+1}\left[t\right].
\end{align*}
and $\beta_{l}\left(\cdot\right)$ is a known Gaussian density. The
density $p_{S}^{k+1|k}\left(t,x^{1:i}|h\right)$ can be directly written
as in (\ref{eq:Trajectory_Gaussian_hypothesis}). 

\subsection{Update}

We compute the quantities in Lemma \ref{lem:Update_radar}. We denote
\begin{align*}
p^{k|k-1}\left(X|h_{j}^{k|k-1}\right) & =\mathcal{N}\left(X;t_{h_{j}^{k|k-1}}^{k|k-1},m_{h_{j}^{k|k-1}}^{k|k-1},P_{h_{j}^{k|k-1}}^{k|k-1}\right).
\end{align*}
Function $\psi_{\mathbf{z}^{k}}\left(\cdot\right)$, which is given
by (\ref{eq:Psi_radar}), is only used for the hypotheses with trajectories
that exist up to time step $k$. Therefore, we can write 

\begin{align*}
\psi_{\mathbf{z}^{k}}\left(t,x^{1:i}|\xi_{j}\right) & =\left\{ \begin{array}{cc}
\frac{p_{D}}{\kappa\left(z_{\xi_{j}}^{k}\right)}\mathcal{N}\left(z_{\xi_{j}}^{k};Hx^{i},R\right) & \xi_{j}>0\\
1-p_{D} & \xi_{j}=0,
\end{array}\right.
\end{align*}
where $t+i-1=k$. Then, for the alive trajectories, we have that 
\begin{align*}
p_{U}^{k|k}\left(X|h_{j}^{k|k-1},\xi_{j}\right) & =\mathcal{N}\left(X;t_{h_{j}^{k|k}}^{k|k},m_{h_{j}^{k|k}}^{k|k},P_{h_{j}^{k|k}}^{k|k}\right)
\end{align*}
where $t_{h_{j}^{k|k}}^{k|k}=t_{h_{j}^{k|k-1}}^{k|k-1}$ and we proceed
to explain how to obtain the other two terms. For $\xi_{j}=0$, we
have $m_{h_{j}^{k|k}}^{k|k}=m_{h_{j}^{k|k-1}}^{k|k-1}$ and $P_{h_{j}^{k|k}}^{k|k}=P_{h_{j}^{k|k-1}}^{k|k-1}$.
For $\xi_{j}\neq0$, we apply the Kalman filter update \cite{Sarkka_book13}
on the trajectory
\begin{align*}
m_{h_{j}^{k|k}}^{k|k} & =m_{h_{j}^{k|k-1}}^{k|k-1}+P_{h_{j}^{k|k-1}}^{k|k-1}\dot{H}^{T}S^{-1}\left(z_{\xi_{j}}^{k}-\overline{z}_{\xi_{j}}^{k}\right)\\
P_{h_{j}^{k|k}}^{k|k} & =P_{h_{j}^{k|k-1}}^{k|k-1}-P_{h_{j}^{k|k-1}}^{k|k-1}\dot{H}^{T}S^{-1}\dot{H}P_{h_{j}^{k|k-1}}^{k|k-1}\\
\overline{z}_{\xi_{j}}^{k} & =\dot{H}m_{h_{j}^{k|k-1}}^{k|k-1},\:S=\dot{H}P_{h_{j}^{k|k-1}}^{k|k-1}\dot{H}^{T}+R\\
\dot{H} & =\left[0_{1,i_{h_{j}^{k|k-1}}-1},1\right]\otimes H
\end{align*}
where $i_{h_{j}^{k|k-1}}$ is the length of the trajectory in hypothesis
$h_{j}^{k|k-1}$. Finally, the normalising constant is
\begin{align*}
\eta_{\mathbf{z}^{k}}\left(h_{j}^{k|k-1},\xi_{j}\right) & =\left\{ \begin{array}{cc}
\frac{p_{D}}{\kappa\left(z_{\xi_{j}}^{k}\right)}\mathcal{N}\left(z_{\xi_{j}}^{k};\overline{z}_{\xi_{j}}^{k},S\right) & \xi_{j}>0\\
1-p_{D} & \xi_{j}=0.
\end{array}\right.
\end{align*}

\section{\label{sec:Appendix_marginalisatioin}}

Here, we prove Theorem \ref{thm:Marginalisation}. If we have a multitrajectory
density $\pi\left(\cdot\right)$ and a transition kernel $f\left(\cdot|\cdot\right)$,
the multitrajectory density $\pi'\left(\cdot\right)$ of the transitioned
set of trajectories is \cite{Mahler_book14}
\begin{align*}
\pi'\left(\mathbf{Y}\right) & =\int f\left(\mathbf{Y}|\mathbf{X}\right)\pi\left(\mathbf{X}\right)\delta\mathbf{X}
\end{align*}
which is analogous to the prediction step (\ref{eq:prediction_trajectories}).
An RFS of targets is an RFS of trajectories whose lengths are one
and the start time information has been discarded. Consequently, the
same type of relation holds if the transition is for RFS of targets,
\begin{align*}
\pi'\left(\mathbf{y}\right) & =\int f\left(\mathbf{y}|\mathbf{X}\right)\pi\left(\mathbf{X}\right)\delta\mathbf{X}.
\end{align*}
In Theorem \ref{thm:Marginalisation}, the transition is $\mathbf{y}=\tau^{k}\left(\mathbf{X}\right)$.
Therefore, it is represented by a multitarget Dirac delta $f\left(\mathbf{y}|\mathbf{X}\right)=\delta_{\tau^{k}\left(\mathbf{X}\right)}\left(\mathbf{y}\right)$
and we obtain Theorem \ref{thm:Marginalisation}. 

\section{\label{sec:Appendix_relation_usual_RFS}}

This appendix shows the relations in Figure \ref{fig:Relation-marginalisation-prediction-update}.
The update step is trivial so we focus on the prediction. Theorem
\ref{thm:Marginalisation} can be written as
\begin{align}
 & \pi_{\tau}^{k}\left(\left\{ z_{1},...,z_{m}\right\} \right)\nonumber \\
 & \quad=\sum_{n=m}^{\infty}\frac{1}{n!}\int\delta_{\tau^{k}\left(\left\{ X_{1},...,X_{n}\right\} \right)}\left(\left\{ z_{1},...,z_{m}\right\} \right)\nonumber \\
 & \quad\quad\times\pi\left(\left\{ X_{1},...,X_{n}\right\} \right)dX_{1:n}\nonumber \\
 & \quad=m!\sum_{n=m}^{\infty}\frac{1}{n!}c_{m}^{n}\int\left[\prod_{j=1}^{m}\delta_{\tau^{k}\left(X_{j}\right)}\left(\left\{ z_{j}\right\} \right)\right]\nonumber \\
 & \qquad\times\left[\prod_{j=m+1}^{n}\delta_{\tau^{k}\left(X_{j}\right)}\left(\emptyset\right)\right]\pi\left(\left\{ X_{1},...,X_{n}\right\} \right)dX_{1:n}.\label{eq:appendix_marginalisation}
\end{align}

\subsubsection*{Prediction without births (targets)}

The prior multitarget density without target births is \cite[Sec 13.2.3]{Mahler_book07}
\begin{align}
 & \pi_{S,\tau}^{k|k-1}\left(\left\{ z_{1},...,z_{m}\right\} \right)\nonumber \\
 & \quad=m!\sum_{n=m}^{\infty}\frac{1}{n!}c_{m}^{n}\int\left[\prod_{i=1}^{n}\left(1-p_{S}\left(y_{i}\right)\right)\right]\nonumber \\
 & \quad\quad\times\left[\prod_{i=1}^{m}\frac{p_{S}\left(y_{i}\right)g\left(z_{i}\left|y_{i}\right.\right)}{1-p_{S}\left(y_{i}\right)}\right]\pi_{\tau}^{k-1}\left(\left\{ y_{1},...,y_{n}\right\} \right)dy_{1:n}\label{eq:prediction_no_births_targetRFS}
\end{align}
where $m!c_{n,m}$ indicates the total number of associations from
$n$ elements to $m$ elements.

\subsubsection*{Prediction without births (trajectories) }

We compute the multitarget density $a_{\tau}^{k}\left(\cdot\right)$
at time $k$ from the predicted multitrajectory density without target
births $\pi_{S}^{k|k-1}\left(\cdot\right)$, via Theorem \ref{thm:Marginalisation},
and show that is equal to $\pi_{S,\tau}^{k|k-1}\left(\cdot\right)$,
see Eq. (\ref{eq:prediction_no_births_targetRFS}). Using (\ref{eq:appendix_marginalisation}),
we get
\begin{align*}
 & a_{\tau}^{k}\left(\left\{ z_{1},...,z_{m}\right\} \right)\\
 & =m!\sum_{n=m}^{\infty}\frac{1}{n!}c_{m}^{n}\int\left[\prod_{j=1}^{m}\delta_{\tau^{k}\left(X_{j}\right)}\left(\left\{ z_{j}\right\} \right)\right]\\
 & \quad\times\left[\prod_{j=m+1}^{n}\delta_{\tau^{k}\left(X_{j}\right)}\left(\emptyset\right)\right]\pi_{S}^{k|k-1}\left(\left\{ X_{1},...,X_{n}\right\} \right)dX_{1:n}\\
 & =m!\sum_{n=m}^{\infty}\frac{1}{n!}c_{m}^{n}\int\left[\prod_{j=1}^{m}\left(1-\delta_{\tau^{k-1}\left(X_{j}\right)}\left(\emptyset\right)\right)\right]\\
 & \quad\times\left[\prod_{j=1}^{n}\left(1-p_{S}\left(X_{j}^{|}\right)\right)^{\left|\tau^{k-1}\left(X_{j}\right)\right|}\right]\\
 & \quad\times\left[\prod_{j=1}^{m}\frac{g\left(z_{j}\left|X_{j}^{|}\right.\right)p_{S}\left(X_{j}^{|}\right)}{1-p_{S}\left(X_{j}^{|}\right)}\right]\pi^{k-1}\left(\left\{ X_{1},...,X_{n}\right\} \right)dX_{1:n}
\end{align*}
where we have used Theorem \ref{thm:General_prediction} and $X_{j}^{|}$
denotes the last target state of trajectory $X_{j}$. Let $p$ denote
the number of targets present at time $k-1$, which satisfies $m\leq p\leq n$.
Then, there are $p-m$ targets present at time $k-1$ that are not
present at time $k$. We can write the previous equation as
\begin{align*}
 & a_{\tau}^{k}\left(\left\{ z_{1},...,z_{m}\right\} \right)\\
 & =m!\sum_{n=m}^{\infty}\sum_{p=m}^{n}\frac{1}{n!}c_{m}^{n}c_{p-m}^{n-m}\int\left[\prod_{j=1}^{p}\left(1-\delta_{\tau^{k-1}\left(X_{j}\right)}\left(\emptyset\right)\right)\right]\\
 & \quad\times\left[\prod_{j=p+1}^{n}\delta_{\tau^{k-1}\left(X_{j}\right)}\left(\emptyset\right)\right]\left[\prod_{j=1}^{p}\left(1-p_{S}\left(X_{j}^{|}\right)\right)\right]\\
 & \quad\times\left[\prod_{j=1}^{m}\frac{g\left(z_{j}\left|X_{j}^{|}\right.\right)p_{S}\left(X_{j}^{|}\right)}{1-p_{S}\left(X_{j}^{|}\right)}\right]\pi^{k-1}\left(\left\{ X_{1},...,X_{n}\right\} \right)dX_{1:n}\\
 & =m!\sum_{p=m}^{\infty}\sum_{n=p}^{\infty}\frac{1}{n!}c_{p}^{n}c_{m}^{p}\int\int\left[\prod_{j=1}^{p}\delta_{\tau^{k-1}\left(X_{j}\right)}\left(\left\{ y_{j}\right\} \right)\right]\\
 & \quad\times\left[\prod_{j=p+1}^{n}\delta_{\tau^{k-1}\left(X_{j}\right)}\left(\emptyset\right)\right]\left[\prod_{j=1}^{p}\left(1-p_{S}\left(y_{j}\right)\right)\right]\\
 & \quad\times\left[\prod_{j=1}^{m}\frac{g\left(z_{j}\left|y_{j}\right.\right)p_{S}\left(y_{j}\right)}{1-p_{S}\left(y_{j}\right)}\right]\pi^{k-1}\left(\left\{ X_{1},...,X_{n}\right\} \right)dX_{1:n}dy_{1:p}\\
 & =m!\sum_{p=m}^{\infty}\frac{1}{p!}c_{m}^{p}\int\left[\prod_{j=1}^{p}\left(1-p_{S}\left(y_{j}\right)\right)\right]\\
 & \quad\times\left[\prod_{j=1}^{m}\frac{g\left(z_{j}\left|y_{j}\right.\right)p_{S}\left(y_{j}\right)}{1-p_{S}\left(y_{j}\right)}\right]\pi_{\tau}^{k-1}\left(\left\{ y_{1},...,y_{p}\right\} \right)dy_{1:p}\\
 & =\pi_{S,\tau}^{k|k-1}\left(\left\{ z_{1},...,z_{m}\right\} \right).
\end{align*}

\subsubsection*{Prediction with births (targets)}

Using the convolution formula \cite{Mahler_book07}, the prior at
time $k$ is
\begin{align}
\pi_{\tau}^{k|k-1}\left(\mathbf{y}\right)= & \sum_{\mathbf{w}\subseteq\mathbf{y}}\pi_{S,\tau}^{k|k-1}\left(\mathbf{w}\right)\beta_{\tau}\left(\mathbf{y}\setminus\mathbf{w}\right).\label{eq:appendix_new_born_targets_conv}
\end{align}

\subsubsection*{Prediction with births (trajectories)}

Using Theorem \ref{thm:Marginalisation}, 
\begin{align*}
\pi_{S,\tau}^{k|k-1}\left(\mathbf{w}\right) & =\int\delta_{\tau^{k}\left(\mathbf{X}\right)}\left(\mathbf{w}\right)\pi_{S}^{k|k-1}\left(\mathbf{X}\right)\delta\mathbf{X}\\
\beta_{\tau}\left(\mathbf{y}\right) & =\int\delta_{\tau^{k}\left(\mathbf{X}\right)}\left(\mathbf{y}\right)\beta^{k}\left(\mathbf{X}\right)\delta\mathbf{X}.
\end{align*}

We have that for RFS of trajectories 
\begin{align*}
\pi^{k|k-1}\left(\mathbf{Y}\right) & =\sum_{\mathbf{W}\subseteq\mathbf{Y}}\pi_{S}^{k|k-1}\left(\mathbf{Y}\setminus\mathbf{W}\right)\beta^{k}\left(\mathbf{W}\right)\\
 & =\pi_{S}^{k|k-1}\left(\mathbf{Y}\cap\mathbb{S}'\right)\beta^{k}\left(\mathbf{Y}\cap\mathbb{B}'\right)
\end{align*}
where $\mathbb{S}'$ and $\mathbb{B}'$ are the disjoint spaces of
sets of trajectories that survive at time $k$ and are born at time
$k$, respectively.

By applying the convolution formula to multitarget Dirac deltas, it
is met that
\begin{align*}
\delta_{\mathbf{a}\uplus\mathbf{b}}\left(\mathbf{y}\right) & =\sum_{\mathbf{w}\subseteq\mathbf{y}}\delta_{\mathbf{a}}\left(\mathbf{w}\right)\delta_{\mathbf{b}}\left(\mathbf{y}\setminus\mathbf{w}\right).
\end{align*}
The marginal density of $\pi^{k|k-1}\left(\cdot\right)$ is denoted
as $b_{\tau}\left(\cdot\right)$. In the following, we show that it
is equal to $\pi_{\tau}^{k|k-1}\left(\cdot\right)$, see (\ref{eq:appendix_new_born_targets_conv}).
Using Theorem \ref{thm:Marginalisation},
\begin{align*}
b_{\tau}\left(\mathbf{y}\right) & =\int\delta_{\tau^{k}\left(\mathbf{X}\right)}\left(\mathbf{y}\right)\pi^{k|k-1}\left(\mathbf{X}\right)\delta\mathbf{X}\\
 & =\int\delta_{\tau^{k}\left(\mathbf{X}\right)}\left(\mathbf{y}\right)\pi_{S}^{k|k-1}\left(\mathbf{\mathbf{X}}\cap\mathbb{S}'\right)\beta^{k}\left(\mathbf{X}\cap\mathbb{B}'\right)\delta\mathbf{X}\\
 & =\int\sum_{\mathbf{w}\subseteq\mathbf{y}}\delta_{\tau^{k}\left(\mathbf{X}\cap\mathbb{S}'\right)}\left(\mathbf{w}\right)\delta_{\tau^{k}\left(\mathbf{X}\cap\mathbb{B}'\right)}\left(\mathbf{y}\setminus\mathbf{w}\right)\\
 & \quad\times\pi_{S}^{k|k-1}\left(\mathbf{\mathbf{X}}\cap\mathbb{S}'\right)\beta^{k}\left(\mathbf{X}\cap\mathbb{B}'\right)\delta\mathbf{X}\\
 & =\sum_{\mathbf{w}\subseteq\mathbf{y}}\int\delta_{\tau^{k}\left(\mathbf{X}\cap\mathbb{S}'\right)}\left(\mathbf{w}\right)\pi_{S}^{k|k-1}\left(\mathbf{\mathbf{X}}\cap\mathbb{S}'\right)\\
 & \quad\times\delta_{\tau^{k}\left(\mathbf{X}\cap\mathbb{B}'\right)}\left(\mathbf{y}\setminus\mathbf{w}\right)\beta^{k}\left(\mathbf{X}\cap\mathbb{B}'\right)\delta\mathbf{X}.
\end{align*}

Using the properties of the set integrals in joint spaces \cite[Sec. 3.5.3]{Mahler_book14},
if $\mathbf{X}=\mathbf{X}_{1}\uplus\mathbf{X}_{2}$, then 
\begin{align*}
\int f\left(\mathbf{X}_{1}\right)g\left(\mathbf{X}_{2}\right)\delta\mathbf{X} & =\int f\left(\mathbf{X}_{1}\right)\delta\mathbf{X}_{1}\int g\left(\mathbf{X}_{2}\right)\delta\mathbf{X}_{2}.
\end{align*}
Therefore, 
\begin{align*}
b_{\tau}\left(\mathbf{y}\right) & =\sum_{\mathbf{w}\subseteq\mathbf{y}}\left[\int\delta_{\tau^{k}\left(\mathbf{X}\right)}\left(\mathbf{w}\right)\pi_{S}^{k|k-1}\left(\mathbf{\mathbf{X}}\right)\delta\mathbf{X}\right]\\
 & \quad\times\left[\int\delta_{\tau^{k}\left(\mathbf{X}\right)}\left(\mathbf{y}\setminus\mathbf{w}\right)\beta^{k}\left(\mathbf{\mathbf{X}}\right)\delta\mathbf{X}\right]\\
 & =\sum_{\mathbf{w}\subseteq\mathbf{y}}\pi_{S,\tau}^{k|k-1}\left(\mathbf{w}\right)\beta_{\tau}\left(\mathbf{y}\setminus\mathbf{w}\right)
\end{align*}
which finishes the proof by comparison with (\ref{eq:appendix_new_born_targets_conv}).

\bibliographystyle{IEEEtran}
\bibliography{21F__Trabajo_laptop_Mis_articulos_Finished_Set_of_trajectories_framework_Accepted_Referencias}

\end{document}